# Narrative Planning: Balancing Plot and Character

**Mark O. Riedl**                                                    RIEDL@CC.GATECH.EDU
*School of Interactive Computing*
*Georgia Institute of Technology*
*Atlanta, GA 30332 USA*

**R. Michael Young**                                                 YOUNG@CSC.NCSU.EDU
*Department of Computer Science*
*North Carolina State University*
*Raleigh, NC 27695 USA*

## Abstract

Narrative, and in particular storytelling, is an important part of the human experience. Consequently, computational systems that can reason about narrative can be more effective communicators, entertainers, educators, and trainers. One of the central challenges in computational narrative reasoning is *narrative generation*, the automated creation of meaningful event sequences. There are many factors – logical and aesthetic – that contribute to the success of a narrative artifact. Central to this success is its *understandability*. We argue that the following two attributes of narratives are universal: (a) the logical causal progression of plot, and (b) character believability. Character believability is the perception by the audience that the actions performed by characters do not negatively impact the audience's suspension of disbelief. Specifically, characters must be perceived by the audience to be intentional agents. In this article, we explore the use of refinement search as a technique for solving the narrative generation problem – to find a *sound and believable* sequence of character actions that transforms an initial world state into a world state in which goal propositions hold. We describe a novel refinement search planning algorithm – the Intent-based Partial Order Causal Link (IPOCL) planner – that, in addition to creating causally sound plot progression, reasons about character intentionality by identifying possible character goals that explain their actions and creating plan structures that explain why those characters commit to their goals. We present the results of an empirical evaluation that demonstrates that narrative plans generated by the IPOCL algorithm support audience comprehension of character intentions better than plans generated by conventional partial-order planners.

## 1. Introduction

Narrative as entertainment, in the form of oral, written, or visual storytelling, plays a central role in many forms of entertainment media, including novels, movies, television, and theatre. Narrative is also used in education and training contexts to motivate and to illustrate. One of the reasons for the prevalence of storytelling in human culture may be due to the way in which narrative is a cognitive tool for situated understanding (Bruner, 1990; McKoon & Ratcliff, 1992; Gerrig, 1993, 1994; Graesser, Singer, & Trabasso, 1994). There is evidence that suggests that we, as humans, build cognitive structures that represent the real events in our lives using models similar to the ones used for narrative in order to better understand the world around us (Bruner, 1990). This *narrative intelligence* (Blair & Meyer, 1997; Mateas & Sengers, 1999) is central in the cognitive processes that we employ across a range of experiences, from entertainment contexts to active learning.

Computational systems that reason about narrative intelligence are able to interact with human users in a natural way because they understand collaborative contexts as emerging narrative and are able to express themselves through storytelling. The standard approach to incorporating storytelling into a computer system, however, is to script a story at design time and then to have the story's script execute without variation at run-time. For a computer system to use a scripted story means that the ability of the system to adapt to the user's preferences and abilities is limited. The story scripted into a system may not completely engage the user's interests or may be too challenging for the user to follow. Furthermore, if stories are scripted at design time, a system can only have a limited number of stories it can present to the user. In entertainment applications, a limited number of stories or a limited number of permutations of a single story results in limited opportunities for user interaction (or limited *replay value* if the computational system is a computer game). In educational and training applications, a limited number of stories or a limited number of permutations of a single story limits the ability of the system to adapt to a learner's needs and abilities.

An alternative approach is to generate stories either dynamically or on a per-session basis (one story per time the system is engaged). Narrative generation is a process that involves the selection of narrative content (the events that will be presented to an audience), ordering of narrative content, and presentation of narrative content through discourse. A system that can generate stories is capable of adapting narrative to the user's preferences and abilities, has expanded replay value, and is capable of interacting with the user in ways that were not initially envisioned by system designers. While many entertainment, educational, and training systems incorporate aspects of storytelling, very few systems exist that generate novel narrative content in order to support the particular needs and preferences of the user. The ability to customize narrative content to the user is the primary motivation of the research effort described in this article.

Narrative content must be understandable, regardless of the purpose of the system that utilizes a narrative generator and the needs of the system user. Of the many factors – both logical and aesthetic – that relate to narrative understandability, we focus on two attributes of narratives we consider to be relatively universal: (a) the *logical causal progression* of plot and (b) *character believability*. Logical progression of plot refers to a property of narrative in which the central events of the narrative obey the rules of the world in which the narrative occurs. Character believability (Bates, 1994) is the perception by the audience that the actions performed by characters do not negatively impact the audience's suspension of disbelief. Specifically, characters must be perceived by the audience to be intentional agents (Dennett, 1989). Thus a believable narrative sequence is one in which all characters can be perceived to be intentional agents.

In this article we describe a narrative generation system that models the fictional narrative creation process as a search-based planning process. The resulting artifact – the plan – is a description of the temporally ordered sequence of actions that story world characters will perform. This plan, when executed or rendered into natural language, tells a story. Plans have been found to be good computational representations of narratives because plans encode attributes central to narrative: action, temporality, and causality (Young, 1999). Unfortunately, solving the planning problem does not also solve the narrative generation problem because planners do not consider many of the logical and aesthetic properties of narratives. Specifically, planners do not consider character believability. We describe a novel refinement search planner – the Intent-based Partial Order Causal Link (IPOCL) planner – that, in addition to creating causally sound plot progression, reasons about character intentionality by (a) identifying possible character goals that explain their actions and (b) creating plan structures that explain why those characters commit to their goals. We begin with a brief background on narrative and lay the theoretical groundwork for planning-based narrative generation (Section 2). Section 3 discusses related work in narrative generation. In Section 4, we lay out our algorithm, IPOCL, for narrative planning in detail and illustrate its processing through examples. Finally, in Section 5, we describe how we evaluated the system.

## 2. Narrative and Planning

In this section we cover some of the relevant background on narrative from the humanities and from cognitive psychology. We use the introduced concepts related to narrative to build an argument for using planning technologies to generate narratives and why off-the-shelf planners, with their emphasis on goal satisfaction, are insufficient.

### 2.1 Narrative Background

Narrative and storytelling are terms that are widely understood but not often well defined. One definition is given here:

**Narrative:** A narrative is the recounting of a sequence of events that have a continuant subject and constitute a whole (Prince, 1987).

For a narrative to have a continuant subject and constitute a whole, the events described in the narrative have a single point or relate to a single communicative goal (Chatman, 1993). One can, however distinguish between narratives that tell a story and narratives that do not (Herman, 2002). A narrative that tells a story has certain properties that one comes to expect. In particular, a story is a narrative that has a plot – the outline of main incidents in a narrative – that is structured to have a particular effect on the audience over time.

Narratologists break narrative down into two layers of interpretation: *fabula* and *sjuzet* (Bal, 1998). The fabula of a narrative is an enumeration of all the events that occur in the story world between the time the story begins and the time the story ends. The events in the fabula are temporally sequenced in the order that they occur, which is not necessarily the same order in which they are told. The sjuzet of a narrative is a subset of the fabula that is presented via narration to the audience. If the narrative is written or spoken word, the narration is in natural language. If the narrative is a cinematic presentation, the narration is through the actions of actors and the camera shots that capture that action. While it is the narrated sjuzet that is directly exposed to the audience, it is the fabula of a narrative that is the content of the narrative, what the narrative is about. In this article, our work is primarily concerned with the generation of a fabula. We assume that a sjuzet can be generated from a fabula in a distinct process (e.g., Callaway & Lester, 2002; Young, 2006; Jhala, 2009; Bae & Young, 2008; Cheong & Young, 2008).

There are many aspects that determine whether a story is accepted by the audience as good. Many of these aspects are subjective in nature, such as the degree to which the audience empathizes with the protagonist. Other aspects appear to be more universal across a wide variety of genres. Cognitive psychologists have determined that the ability of an audience to comprehend a narrative is strongly correlated with the causal structure of the story (Trabasso & Sperry, 1985; van den Broek, 1988; Graesser, Lang, & Roberts, 1991; Graesser et al., 1994) and the attribution of intentions to the characters that are participants in the events (Graesser et al., 1991; Gerrig, 1993; Graesser et al., 1994). Story comprehension requires the audience (e.g. reader, hearer, viewer) to perceive the causal connectedness of story events and to infer intentionality of characters. Accordingly, the two attributes of narrative that we focus on in this work on narrative generation are *logical causal progression* and *character believability*.

The causality of events is an inherent property of narratives and ensures a whole and continuant subject (Chatman, 1993). Causality refers to the notion that there is a relationship between temporally ordered events such that one event changes the story world in a particular way that enables future events to occur (Trabasso & van den Broek, 1985). For a story to be considered successful, it must contain a degree of causal coherence that allows the audience to follow the logical succession of events and predict possible outcomes. Attesting to the importance of causality in story, Trabasso and Sperry (1985) found a significant correlation between recall of an event in a story and its existence as part of a causal chain that terminates in the outcome of the story.

Character believability (Bates, 1994) is the perception by the audience that the actions performed by characters do not negatively impact the audience's suspension of disbelief. Character believability is partially dependent on the idiosyncrasies of a character's appearance and physical movements. Physical appearance is very important in visual media such as animated film (Thomas & Johnson, 1981). Descriptions of character appearances are also found in written and spoken presentations. Equally important is the way in which the internal attributes of a character such as personality, emotion, desires, and intentions manifest themselves through the decisions the character makes and the behaviors the character performs (Thomas & Johnson, 1981; Bates, 1994; Loyall, 1997).[1] The definition of character believability places emphasis on the goal-oriented nature of characters. Goal-oriented behavior is a primary requirement for believability (Loyall, 1997; Charles, Lozano, Mead, Bisquerra, & Cavazza, 2003). Specifically, we, as humans, ascribe intentionality to agents with minds (Dennett, 1989). The implication is that if a character is to be perceived as believable, one should be able to, through observations of the character, infer and predict its motivations and intentions. In this article, our approach to narrative generation focuses explicitly on creating narrative sequences in which characters will be perceived to be intentional agents. Other research efforts have directly addressed other aspects of character believability, including personality (e.g., Carbonell, 1980; Reilly, 1996; Rizzo, Veloso, Miceli, & Cesta, 1999; Sengers, 2000), emotion (e.g., Gratch & Marsella, 2004; Seif El-Nasr, Yen, & Ioerger, 2000), and appearance and physical performance (e.g., Blumberg & Galyean, 1995; Maes, Darrell, Blumberg, & Pentland, 1995; Perlin & Goldberg, 1996; Loyall, 1997; Hayes-Roth, van Gent, & Huber, 1997; Lester, Voerman, Towns, & Callaway, 1999).

---

1. Loyall (1997) enumerates many of the elements that affect character believability in autonomous agents.

## 2.2 Planning as a Model of Narrative Generation

There are many parallels between plans and narrative at the level of fabula. In particular, a narrative is a sequence of events that describes how the story world changes over time. In a fabula, change is instigated by intentional actions of story world characters, although the story world can also be changed through unintentional acts such as accidents and forces of nature. Likewise, a plan is a set of ordered operators that transforms a world from one state to another state. If the operators of a plan are events that can happen in a story world, then a plan can be a model of a fabula. *Partially ordered* plans allow operations to remain temporally unconstrained if their relative execution order does not matter. The semantics of the plan and the capabilities of the plan execution engine may determine whether operations can, in fact, be executed in parallel (Knoblock, 1994). Similarly, the events in a fabula can occur simultaneously in the story world, even though the narration (e.g., sjuzet) of the events is necessarily linear.

Planners are implementations of algorithms that solve the planning problem: given a domain theory, an initial state $I$, and a goal situation $G$ consisting of a set of propositions, find a sound sequence of actions that maps the initial state into a state where $G$ is true. The *domain theory* is a model of how the world can change. For example, one can use STRIPS (Fikes & Nilsson, 1971) or STRIPS-like operators that specify what operations can be performed in the world, when they are applicable, and how the world is different afterward. Various algorithms have been developed that solve planning problems including partial-order planners, constraint satisfaction planners, and heuristic search planners.

Since a plan can be used as a model of fabula, a planning algorithm can also be used as a model of the dramatic authoring process that humans use to create narratives. Thus, the creation of a narrative can be considered a problem solving activity if one considers the fabula of a narrative to be the sequence of story-world events that achieves some outcome desired by the author in order to have some effect or impact on an audience.

In this article, we present an algorithm for planning narratives. It specifically solves the fabula planning problem.

**Fabula Planning Problem:** Given a domain theory, find a *sound* and *believable* sequence of character actions that transforms an initial world state $I$ into a world state in which goal propositions $G$ hold.

The domain theory, initial state, and goal situation are provided by the user of the fabula generation system, whom we call the *human author*. The fabula generation system is tasked with selecting and ordering a set of actions that, when told (as opposed to executed), is considered a narrative.

The algorithm presented in subsequent sections can be considered one example of an algorithm that solves the fabula generation problem. As with planning algorithms in general, we acknowledge that other algorithms may exist. In the next sections, we explore the implications of searching for *believable* narrative plans.

### 2.2.1 CHALLENGES OF PLANNING AS A COMPUTATIONAL MODEL OF NARRATIVE GENERATION

Algorithms that solve the planning problem find sequences of operations that are sound, meaning that, in the absence of non-determinism, they are guaranteed to find a sequence of operations that maps the initial state into a state in which the goal situation holds. When generating a fabula, we assume that operations are actions to be performed by characters that exist in a story world. A consequence of the planning problem definition is that planners do not consider whether it is *natural* or *believable* for a character to perform an action at any given time during the plan; they do not consider actions from the perspective of the character or the audience, but from the context of whether it is necessary for the goal situation to be achieved. We argue that this limits the applicability of off-the-shelf planners as techniques for generating stories.

To illustrate this limitation, we present a simple example. Suppose we describe a world with three characters: a king, a knight, and a princess. All characters live in a castle and the castle has a tower in which characters can be locked up. Further suppose a goal situation has been provided by the human author in which the princess is locked in the tower and the king is dead. Given a reasonable domain theory – e.g., a set of possible actions – one plan that can be found by a planner is:

1. The princess kills the king.

2. The princess locks herself in the tower.

This plan is valid from the perspective that it transforms the initial state into a state in which the goal situation holds. But does it make sense as a story? A reader of this short story will be left with many questions in mind. Why does the princess kill the king? Why does the princess lock herself in the tower? Other plans exist that might make more sense, such as:

1. The king locks the princess in the tower.

2. The knight kills the king.

Intuitively, it is easier for the reader to find an explanation that makes sense of this second story: the princess must have upset the king and the knight must be avenging the princess. Let us consider ways we can influence a planner to give us more favorable plans.

One possibility is that we could modify the problem definition. For example, we can change the initial state and domain theory such that princesses cannot kill kings or that characters cannot lock themselves in the tower. Do these modifications make sense? Suppose the king were attempting to harm the princess – the princess would be justified in killing the king. To declare that princesses can never kill kings seems to impose an unnecessarily strong assumption on what narratives can be generated. Likewise, we can imagine narratives in which it makes sense to lock oneself in a tower (perhaps to escape from danger). One of the advantages of automated generation of narratives is that an algorithm can explore many possibilities and/or create narratives that were not envisioned by the human author. This argument is further expanded by Riedl and Young (2006) in the context of creativity on the part of a computational system.

Another possibility is that we provide a heuristic function to a planner that favorably ranks plans that demonstrate the quality of character believability. Such a heuristic will increase the probability that the planner find a solution that is believable by ranking plans as being "closer" to being a solution when they include actions that create the appearance of intentionality. For example, a planner with a good heuristic could, in principal, find the following plan:

1. The princess and the knight fall in love.

2. The king proposes to the princess.

3. The princess refuses the king's proposal.

4. The king locks the princess in the tower.

5. The knight kills the king.

Like the previous example, this plan has the king lock the princess in the tower and then has the knight kill the king. The inclusion of actions 1 and 3, however, increase the likelihood that a reader will find this story believable; the princess's refusal explains why the king locks the princess in the tower (the king's proposal establishes the conditions for the princess's refusal), and the princess and the knight falling in love explains why princess refuses the king's proposal and why the knight kills the king.

A good heuristic that ranks on believability, however, only increases the probability that a complete plan is found that has the desired properties by making it cost less to explore the portions of the search space in which those solutions are likely to exist. It is still possible for the planner to return a plan that is not believable in situations where it finds a shorter, complete solution before it finds the longer, complete, and believable solution. This occurs because the planning problem is solved when a sound sequence of actions that transforms the initial state into one in which the goal situation holds.

We conclude that the fabula generation problem is sufficiently different than the planning problem that if we wish to automatically plan the actions in a fabula, we can benefit from new definitions for plan completeness and mechanisms for selecting actions that move the planner toward complete solutions. We consider in greater detail what it means for a character to appear believable with respect to intentionality, how we can detect character believability in fabula plans, and how a planner can select actions that directly address both logical causal progression and believability.

### 2.2.2 INTENTIONALITY IN CHARACTER BELIEVABILITY

Character believability is partially due to character intentionality in that a story is more likely to be considered believable if the story world characters appear to be motivated by individual goals and desires. Stories are likely to be found more comprehensible when there are well-formed relationships between character actions and recognizable character goals (Graesser et al., 1991, 1994). However, unlike causal properties of story, there are no structures in plans or processes in planning algorithms that correspond directly to character intentionality except the goal propositions in the planning problem. However, this is complicated by the fact that stories typically comprise of multiple characters who cannot be assumed to have the same goals as the human author, or to want to achieve any of the goal situation propositions provided by the human author.

The goal situation in fabula planning is thus reinterpreted as the *outcome* of the story. However, all the propositions of the outcome are not necessarily intended by all story world characters. Indeed, it is possible that none of the propositions of the goal situation are intended by any of the story world characters. It is also not necessarily the case that any or all of the story world characters have declared intentions at the beginning of the story (in the initial state). That is, characters may only form intentions as a reaction to conditions in the world or in response to the actions of other characters.

Achieving character intentionality in a fabula planner requires a decoupling of the characters' intentions from the intentions of the human author and from the declaration of the initial state and goal situation. Thus, we distinguish between the *author goals* (Riedl, 2004, 2009) and *character goals*. The author goal is a description of the world that the author would like the fabula generator to achieve. For simplicity, we only consider a single author goal encoded into the outcome, although it is often advantageous for the human author to indicate several intermediate situations through which the fabula should pass as means of providing additional guidance as to what he or she desires in a solution (cf., Riedl, 2009). Character goals, on the other hand, are the goals that characters are perceived to pursue through a portion of the overall fabula. Characters goals may be different from the outcome in the sense that not all characters in a story desire the outcome state or seek to achieve it. Character goals may also be adopted and resolved throughout a story. For example, in the previous section the king appears to develop a goal of punishing the princess after she refuses to marry him. The king's goal neither exists at the beginning of the story nor persists through to the outcome.

Once agent intentions are decoupled from the initial world state and goal situation, the planner must assume responsibility for determining character goals – why the agents are performing the actions in the plan – and motivate those intentions with other actions. Failure to distinguish between author goals and character goals results in the appearance of collusion between characters to achieve the outcome situation when it does not makes sense (e.g., a protagonist and antagonist) and/or to act inconsistently and erratically.[2] Our approach to incorporating character intentionality into narrative planning is described in Section 4.

To illustrate the decoupling of character intentions, let us inspect the example planning problem from Section 2.2.1 in more detail. The goal situation has two propositions: (a) the princess is locked in the tower, and (b) the king is dead. This is an example where the human author's intentions – that a particular outcome is reached – is not likely to be the same as reasonable intentions of story world characters. That is, the princess is unlikely to intend to be locked up and the king is unlikely to intend to be dead. Further, it is not clear that there is any reason why the princess or the knight should intend that the king be dead, although these declarations could be made by the human author at initialization time. It would be reasonable for the reader to see the king appear to fight back against the knight to try to avoid death. But the planning problem is not *adversarial* in the sense that there can be any uncertainty about whether the knight will prevail.[3]

---

2. Sengers (2000) refers to this phenomenon as *agent schizophrenia*.

We conclude that a fabula planner must select actions for characters that achieve the outcome situation and that also create the appearance that the story world characters have intentions that are potentially distinct from the human author's desires. Since characters' intentions are not necessarily provided in advance by the human author, reasonable goals for characters must be *discovered* that explain their behaviors, and fabula structure must be constructed that illustrates the formation of those intentions. After reviewing related work, we will describe one algorithm that meets the requirements for solving the fabula generation problem as laid out earlier.

## 3. Related Work

Narrative generation systems can often be classified as using one of two approaches to fictional fabula content creation. *Simulation-based* narrative generation systems (also referred to as *emergent systems* in Aylett, 1999, 2000) are those that simulate a story world. The simulation approach is to establish a set of characters and a world context. The narrative generation system then progressively determines the actions that the characters should take over time as the situational context evolves. Often simulation-based narrative generation systems employ decentralized, autonomous embodied agents that represent story world characters and react to the evolving world state. *Deliberative* narrative generation systems are those that generate narratives by solving the problem of choosing a sequence of actions – physical, mental, and dialogue – for all story world characters that meet certain constraints and parameters (aesthetic, dramatic, or pedagogical). The narrative is the output of this procedure. The primary distinction to simulation-based approaches is that a deliberative narrative generation system uses a centralized reasoning algorithm – often a planner – to determine the optimal actions for all characters. We limit our discussion of related narrative generation research to how systems produce fabula content.

The simulation-based (emergent) approach to narrative generation is based on the assertion that the best way to generate a narrative is to model the behaviors and decision-making processes of story world characters. Tale-Spin (Meehan, 1976) is a system that generates Aesop's Fables based on moment-to-moment inference about what each character should do. The inference engine is based on theories of common-sense reasoning (Schank & Abelson, 1977). Meehan (1977) notes that in circumstances where character goals are not well chosen or where the facts of the story world do not support the character actions the user intends, generated narratives can be very short and oddly structured (see Meehan, 1977, for examples of "mis-spun" narratives). The Carnegie Mellon University Oz project (Bates, 1992, 1994; Mateas, 1997; Loyall, 1997) uses autonomous, reactive, embodied agents to represent characters in a virtual world. The agents use "shallow and broad" (Bates, Loyall, & Reilly, 1992) decision-making routines to cover a wide repertoire of believable-looking activities. As first proposed by Laurel (1986) and later implemented by Bates and colleagues (Bates, 1992; Kelso, Weyhrauch, & Bates, 1993; Weyhrauch, 1997), a special agent, called a *drama manager* may be necessary to prevent uninteresting and poorly structured narratives from emerging. A drama manager oversees and coordinates character agent behavior in order to coerce interesting and well-structured performances out of the autonomous agents. The I-Storytelling system (Cavazza, Charles, & Mead, 2002) likewise relies on autonomous, reactive agents to represent story world characters. Unlike the Oz project, I-Storytelling system agents use hierarchical task network (HTN) planners to achieve pre-determined goals. Cavazza et al. note that the way in which the virtual world is configured, including the initial position of character agents, and the initial goals of character agents strongly influences the outcome of the story; poor initial configurations may result in uninteresting narratives with no conflict.

---

3. This is an example in which agents be perceived to intentionally strive to avoid the human author's desired outcome. Since the goal situation is the human author's intention, the goal situation must be achieved.

Dehn (1981) asserts that the process of computational story generation must be a process that includes the satisfaction of the intentions of the human author. Deliberative narrative generation systems often consider the process of narrative creation from the perspective of a singular author that has authority over the resulting narrative structure and is working to achieve a narrative sequence that conforms to particular given constraints and parameters. The Universe system (Lebowitz, 1984, 1985, 1987) uses a centralized hierarchical planner to produce open-ended narrative soap-opera episodes that achieve the narratological intentions of the human author. The human author provides a goal situation that describes the outcome of a particular episode and the Universe system's planner finds a sequence of character actions that achieves that goal using hierarchically related task networks describing common activity. In general, hierarchical decomposition requires some form of grammar or rules. Story grammars such as that by Rumelhart (1975) have been criticized as too restrictive (Black & Wilensky, 1979; Wilensky, 1983). Tailor (Smith & Witten, 1991) uses state-space search to plan the actions of the story's protagonist. The protagonist is given a goal to achieve and Tailor searches for a sequence of actions the protagonist can take to achieve the goal. When an antagonist is present, Tailor uses adversarial search.

More recent work on deliberative narrative generation systems has focused on two areas: the role of knowledge, and specialized search algorithms. The Minstrel system (Turner, 1994) implements a model of computational creativity based on adaptation and reuse of existing concepts to create new stories. The Minstrel system uses specialized routines to transform old stories into new stories. México (Pérez y Pérez & Sharples, 2001) implements a model of creative writing (cf., Sharples, 1999) that conceptualizes writing as a cycle of cognitive engagement and reflection. The model employs a combination of case-based reasoning – it probes a database of known existing stories for elements that match current patterns of emotion and tension – and partial-order planning  to ensure coherence between story fragments. The ProtoPropp system (Gervás, Díaz-Agudo, Peinado, & Hervás, 2005) uses a case-based reasoning approach to creating narratives. ProtoPropp encodes examples of Russian folktales from Propp (1968) and functions based on regularities about folktales identified by Propp into an ontology from which new folktales are created by retrieving, adapting, and reusing parts of old folktales. More recently Porteous and Cavazza (2009) have turned to planning narrative structures using a variation of FF (Hoffmann & Nebel, 2001) to find a sequence of events that brings about a goal situation. The planner does not consider character goals independent of the goal situation. Instead their generation algorithm uses *landmarks* – partially ordered sets of first-order logic literals that must be made true throughout the course of a solution – as a means of guiding the planner toward a solution consistent with the human author's vision. Landmarks are author goals (Riedl, 2004, 2009).

The goal of our research is to devise a narrative generation system that generates narratives that exhibit both causal coherence and character intentionality. We favor a deliberative approach to narrative generation because a deliberative approach provides a mechanism for ensuring that character actions are chosen with global structure in mind. Contrast this to simulation-based approaches that choose character actions based on temporally localized information. Deliberative systems avoid problems with logical causal progression because they consider the narrative sequence as a whole. However, those deliberative narrative generation systems that have been designed to date conflate character goals and human author goals without consideration for audience – reader, viewer, etc. – perspective.

Our algorithm, the Intent-Driven Partial Order Causal Link (IPOCL) planner, is an algorithm that solves the fabula generation problem. The IPOCL algorithm is based on a class of planning algorithms called Partial Order Causal Link (POCL) planners, of which UCPOP (Penberthy & Weld, 1992; Weld, 1994) is a well known example. POCL planners represent operators in a plan as STRIPS (Fikes & Nilsson, 1971) or STRIPS-like constructs consisting of the operator name, a precondition – the conditions that must be true in a world for an operator to be executable – and an effect – the conditions in the world that are changed once an operator finishes execution. The precondition and effect consist of zero or more first-order logic literals. Operators may be parameterized with variables that, when bound to ground symbols, allows a single operator schema to represent many possible ground operators. POCL planners use the following definition of a partially ordered plan, using the term "step" to refer to operators that are instantiated into the plan structure:

> **Definition 1 (POCL Plan):** A POCL plan is a tuple $\langle S, B, O, L \rangle$ such that $S$ is a set of plan steps, $B$ is a set of binding constraints on the parameters of the steps in $S$, $O$ is a set of temporal orderings of the form $s_1 < s_2$ where $s_1, s_2 \in S$, and $L$ is a set of causal links of the form $\langle s_1, p, q, s_2 \rangle$ where $s_1, s_2 \in S$ and $p$ is an effect of $s_1$ and $q$ is a precondition of $s_2$ and $p$ unifies with $q$.

Note that we use the term *step* synonymously with *action* and *operator*. This differs from the usage of the term in literature on non-POCL planners such as SATPLAN and GRAPHPLAN.

POCL planners use an iterative process of identifying *flaws* in a plan and revising the plan in a least-commitment manner. A flaw is any reason why a plan cannot be considered a valid solution. An *open condition* flaw occurs when the precondition of a step or the goal situation is not satisfied by the effects of a preceding step or the initial state. A POCL planner solves for the open condition of the step or goal situation by non-deterministically choosing existing steps or instantiating new steps that have effects that unify with the goal conditions. A *causal link* (Penberthy & Weld, 1992) connects two plan steps $s_1$ and $s_2$ via condition $p$, written $s_1 \xrightarrow{p} s_2$, when $s_1$ establishes the condition $p$ in the story world needed by subsequent step $s_2$ in order for step $s_2$ to execute. Causal links are used to record the causal relationships between steps and record the satisfaction of open conditions. A *causal threat* flaw occurs when the effects of one plan step possibly undo the effects of another plan step. Causal threats are resolved by explicitly ordering the conflicting steps. We provide

---

**POCL** $(\langle S, B, O, L \rangle, F, \Lambda)$

The first parameter is a plan. On the initial call to POCL, there are only two steps in $S$ – the dummy initial step whose effect is the initial state and the final step. $F$ is a set of flaws. On the initial call, $F$ contains an open condition flaw for each goal literal in the goal situation. $B = L = \emptyset$. $\Lambda$ is the set of action schemata. Output is a complete plan or $fail$.

I. **Termination.** If $O$ or $B$ is inconsistent, fail. Otherwise, if $F$ is empty, return $\langle S, B, O, L \rangle$.

II. **Plan Refinement.**

    1. **Goal selection.** Select an open condition flaw $f = \langle s_{\text{need}}, p \rangle$ from $F$. Let $F' = F - \{f\}$.

    2. **Operator selection.** Let $s_{\text{add}}$ be a step that adds an effect $e$ that can be unified with $p$ (to create $s_{\text{add}}$, non-deterministically choose a step $s_{\text{old}}$ already in $S$ or instantiate an action schema in $\Lambda$). If no such step exists, backtrack. Otherwise, let $S' = S \cup \{s_{\text{add}}\}$, $O' = O \cup \{s_{\text{add}} < s_{\text{need}}\}$, $B' = B \cup B_{\text{new}}$ where $B_{\text{new}}$ are bindings (e.g., assignments of ground symbols to variables) needed to make $s_{\text{add}}$ add $e$, including the bindings of $s_{\text{add}}$ itself, and $L' = L \cup \{\langle s_{\text{add}}, e, p, s_{\text{need}} \rangle\}$. If $s_{\text{add}} \neq s_{\text{old}}$, add new open condition flaws to $F'$ for every precondition of $s_{\text{add}}$.

    3. **Threat resolution.** A step $s_{\text{threat}}$ threatens a causal link $\langle s_j, e, p, s_k \rangle$ when it occurs between $s_j$ and $s_k$ and it asserts $\neg e$. For every used step $s_{\text{threat}}$ that might threaten a causal link $\langle s_j, e, p, s_k \rangle \in L'$, non-deterministically do one of the following.

       • **Promotion.** If $s_k$ possibly precedes $s_{\text{threat}}$, let $O' = O' \cup \{s_k < s_{\text{threat}}\}$.

       • **Demotion.** If $s_{\text{threat}}$ possibly precedes $s_j$, let $O' = O' \cup \{s_{\text{threat}} < s_j\}$.

       • **Separation.** Let $O' = O' \cup \{s_j < s_{\text{threat}}, s_{\text{threat}} < s_k\}$ and let $B' = B' \cup$ the set of variable constraints needed to ensure that $s_{\text{threat}}$ won't assert $\neg e$.

III. **Recursive invocation.** Call POCL $(\langle S', B', O', L' \rangle, F', \Lambda)$.

---

Figure 1: The POCL algorithm.

the POCL planning algorithm in Figure 1 as a point of comparison for later discussion of our fabula planning algorithm.

In the next section, we introduce our algorithm, the Intent-Driven Partial Order Causal Link (IPOCL) planner, which creates narratives that, from the perspective of a reader, more closely resemble the results of an emergent narrative generation system with regard to character intentionality. Specifically, IPOCL is a modification of existing search-based planning algorithms to support character intentionality independent of author intentions. The goal is to generate narratives through a deliberative process such that characters appear to the audience to form intentions and act to achieve those intentions as if they were simulated. In this way, IPOCL can produce narratives that have both logical causal progression, meaning that they achieve author-indicated outcomes states, and have believable characters.

## 4. Intent-Driven Planning

The definition of character believability in this work is constrained to focus on the perceived intentionality of character behavior in the story world. Perceived intentionality refers to the way in which characters are observed by an audience to have goals and to act to achieve those goals. In the context of computational storytelling systems, it is not sufficient for a character to act intentionally if the audience is not capable of inferring that character's intentions from the circumstances that surround the character in the story world. The audience of a story is not a collection of passive observers. Instead, the audience actively performs mental problem-solving activities to predict what characters will do and how the story will evolve (Gerrig, 1993). It makes sense, therefore, to reason about character intentions and motivations at the time of generation from the perspective of the audience. This will ensure that every character action considered for inclusion in the narrative will appear motivated and intentional.

The Intent-Driven Partial Order Causal Link (IPOCL) planner that generates fabula plans in which characters act intentionally and in which that intentionality is observable. IPOCL extends conventional POCL planning to include an expanded plan representation, definition of plan completeness, and action selection mechanisms that facilitate a fabula planner to search for a solution in which the author's goal is achieved (e.g., the outcome) and all characters appear to act intentionally. Conventionally, planners are means-ends tools for solving problems. When employing a planning system to generate a fabula, the system must produce the actions that make up the plot line of a story, along with a temporal ordering – partial or total – over the execution times of those actions. We make the following observations about the conventional planning problem:

- Plans being generated are created by or for a single agent (Bratman, Israel, and Pollack, 1988) or for a collection of cooperating agents (Grosz & Sidner, 1990).[4]

- The goal situation is intended by one or more of the character agents and all agents intend to execute a plan in support of achieving the goal state.

In order to facilitate the active mental processes of the audience suggested by Gerrig, we observe that solving the fabula planning problem requires the following:

- Plans being generated are created for multiple character agents that are not necessarily cooperating but also not necessarily adversarial.

- The goal situation describes properties of the world that are not necessarily intended by any of the character agents that are to execute the plan.

The goal situation for the conventional planning problem is a partial description of the world state that will be obtained at the end of the plan's execution. In the context of narrative planning, we refer to the goal situation as the *outcome* because it describes how the world must be different after the narrative is completed.

The fabula generation algorithm described in the remainder of this section searches the space of plans in which individual agent goals are potentially distinct from the outcome

---

4. The SharedPlans (Grosz & Sidner, 1990) formalism addresses the situation where more than one agent collaborates to construct a joint plan for achieving some goal. Grosz and Sidner's approach addresses the cases where all agents intend that a joint goal is achieved or where one agent has a goal and "contracts out" part of the task to another agent by communicating its intentions (Grosz & Kraus, 1996). SharedPlans address the coordination of many individual plans into a single joint plan by defining how agent intentions to perform actions and agent intentions that goals and sub-goals be achieved constrain the behaviors of the individual agents working together. The formalism, however, does not address situations where agents have different goals and are cooperating or "contracting out."

```
ACTION ::= ACTION-NAME (VARIABLE*)
           actors:  VARIABLE*
           happening:  BOOLEAN
           constraints:  LITERAL*
           precondition:  LITERAL*
           effect:  LITERAL*
LITERAL := PREDICATE ([VARIABLE | SYMBOL]*)
```

Figure 2: Syntax for action schemata in IPOCL.

and in which agents are not necessarily cooperating. Character agents can either be given intentions as part of the specification of the initial world state or develop them during the course of the plan. The IPOCL planning algorithm accomplishes this by expanding the representation of the plan structure to include information about the intentions of the individual agents. Algorithmically, IPOCL simultaneously searches the space of plans and the space of agent intentions. At any point in the process, agent intentions are ensured to be plausible through the use of a special reasoning process that tests for character intentionality from the perspective of the audience and attempts to revise the plan if the test fails.

### 4.1 Extensions to the Planning Problem Definition Language

The IPOCL planning problem is given in Definition 2.

> **Definition 2 (IPOCL Planning Problem):** An IPOCL planning problem is a tuple, $\langle I, A, G, \Lambda \rangle$, such that $I$ is the initial state, $A$ is a set of symbols that refer to character agents, $G$ is the goal situation, and $\Lambda$ is a set of action schemata.

A significant factor in the IPOCL planning problem is $A$, the set of symbols that refer to character agents in the world. These symbols are handled specially in processes determining character intentionality. We have extended the traditional planning problem definition language to use the character agents in two ways:

- Specification of which actions do not need to be intentional.

- Specification of which parameters of an action refer to the character agents that will be intentionally performing the action.

The syntax for specifying action schemata for IPOCL is given in Figure 2. As with other POCL planners, we use a STRIPS-like representation with preconditions and effects. Additionally, *constraints* are literals that must unify with those in the initial state and act as a filter on applicable parameter bindings.

We distinguish between two types of actions: *happenings* (Prince, 1987) and *non-happenings*. Happenings are actions that can occur without the intention of any character such as accidents, involuntary reactions to stimuli, and forces of nature. Non-happening events must be intended by a character. For clarity, we assume that, unless indicated otherwise, all actions are non-happenings and thus require an actor for whom the action fulfills

---

Action: **slay** (?slayer, ?monster, ?place)
    actors: ?slayer
    constraints: knight(?slayer), monster(?monster), place(?place)
    precondition: at(?slayer, ?place), at(?monster, ?place), alive(?slayer), alive(?monster)
    effect: ¬alive(?monster)

Action: **marry** (?groom, ?bride, ?place)
    actors: ?groom, ?bride
    constraints: male(?groom), female(?bride), place(?place)
    precondition: at(?groom, ?place), at(?bride, ?place), loves(?groom, ?bride),
                  loves(?bride, ?groom), alive(?groom), alive(?bride)
    effect: married(?groom), married(?bride), ¬single(?groom), ¬single(?bride),
        married-to(?groom, ?bride), married-to(?bride, ?groom)

Action: **appear-threatening** (?monster, ?char, ?place)
    actors: ?monster
    happening: t
    constraints: monster(?monster), character(?char), place(?place)
    precondition: at(?monster, ?place), at(?char, ?place), scary(?monster), ?monster≠?char
    effect: intends(?char, ¬alive(?monster))

---

Figure 3: Example IPOCL action schemata.

an intention. If there are actions that can occur in the world without intent (for example, falling down the stairs), they are marked by specifying the *happening* slot as true.

If the action is a non-happening, the *actors* slot specifies which of the parameters refer to symbols representing characters that are acting intentionally to enact the particular action. We say that an action is *to be performed by* character agent *a* when the *actors* slot of the action references *a*. Figure 3 shows three action schemata involving a single intentional actor, multiple intentional actors, and no intentional actor, respectively. The action schema for `Slay(?slayer, ?victim, ?place)` specifies that `?slayer` will refer to the intentional actor. Note the implication that slaying cannot be performed accidentally. The action schema for `Marry(?groom, ?bride, ?place)` specifies two intentional actors, `?groom` and `?bride`. Finally, `Appear-threatening(?monster, ?char, ?place)` indicates that `?char` will appear to be become frightened by a monster. This action does not need to be intentional on the part of the `?monster` or `?char`.

A final note on action definition is the use of the special *intends* predicate, which can only be used in the effect of an action. Semantically the *intends* predicate should be read as meaning "it is reasonable for a character to have the following goal" as a response to the action. Whether the intention is acted upon is determine by whether the proposition is used, as described in the next section. When the narrative is told, there will be no mention of facts that are unused. The *intends* predicate only occurs in action effects; it is not used in action preconditions because that creates a strong commitment to how actions can be used. For example, if `Marry` were to require `intends(?groom, married-to(?groom, ?bride))`, it would preclude stories in which a character wants to not be single (but doesn't necessarily want to be married) and also stories in which a character marries someone as revenge against a third party (the marriage is just one action in a chain leading up to another goal that is none of the action's effects).

## 4.2 Character Intentionality in Fabula Planning

Because a story's audience actively performs problem-solving as the story progresses in order to predict the outcome and the fate of story world characters, a generated story should support these cognitive processes. This means providing narrative structure that gives enough information for the audience to infer the intentionality of character behavior. From the fabula planner's perspective, all character actions should be intentional (or happenings). That is, for every character goal, a portion of the actions in the complete fabula plan describe the actions to be performed by that character to achieve the character goal. We formalize this as follows:

> **Definition 3 (Frame of Commitment):** A frame of commitment is a tuple, $\langle S', P, a, g_a, s_f \rangle$, such that $S'$ is a proper subset of plan steps in a plan $P = \langle S, B, O, L \rangle$, $a$ is a symbolic reference to a character agent such that the character agent is the actor of all steps in $S'$, $g_a$ is a goal that character agent $a$ is pursing by executing the steps in $S'$, and $s_f \in S'$ – referred to as the *final step* – has $g_a$ for one of its effects and all other steps in $S'$ temporally precede $s_f$ in the step ordering $O$ of plan $P$.

The purpose of the frame of commitment is to record a character's internal character goal $g_a$ and the actions of the plan that the character will appear to perform during storytelling to achieve that goal. However, from the perspective of the audience, it is not enough to declare a character as having a goal; in order to make inferences about character intentions and plans, the audience must observe the characters forming and committing to goals. Therefore, each frame of commitment is associated with a condition, $e_g$, of the form `intends(a, ga)`, which indicates that for a character to commit to an internal character goal, $a$ must be in a state where it is reasonable to intend $g_a$. The condition $e_g$ must be established in the world by some plan step that has $e_g$ as an effect. That is, something in the world causes character $a$ to commit to $g_a$. The plan step that causes $e_g$ and consequently causes the frame of commitment is referred to as the *motivating step* for the frame of commitment. The motivating step necessarily temporally precedes all plan steps in the frame of commitment. Informally, the *interval of intentionality* is the set of actions $S'$ that character $a$ will perform to achieve the internal character goal, $g_a$.[5]

Character goals partially describe a world state that the character commits to achieving. Commitments persist through time and a character will remain committed to the goal even though the desired world state is undone (Bratman, 1987). IPOCL does not explicitly represent the release of a commitment except to say that the interval of intentionality is bounded temporally by the set of steps in the interval of intentionality.

---

5. An interval of intentionality roughly equates to the notion of a Full Individual Plan (FIP) in the Shared-Plans formulation (Grosz & Sidner, 1990). A full individual plan is a portion of the larger Full Shared Plan (FSP) that a single agent is responsible for executing. The distinction between a fabula plan and an FSP is that the full fabula plan is not made up of many individual FIPs generated by collaborating planning agents. Instead, a fabula plan is constructed as a whole and the individual character actions that make up the whole plan are annotated to indicate what intention they might be used to achieve.

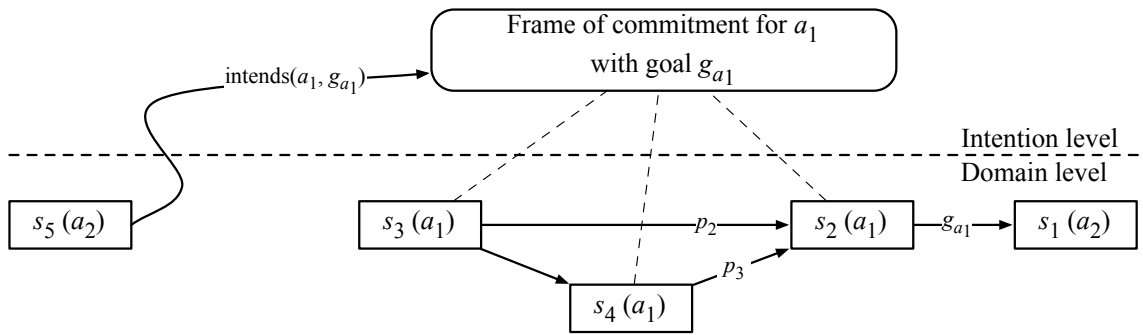

Figure 4: An IPOCL plan with a single frame of commitment and motivating step.

Character goals are captured in two ways. First, potential character intentions are recorded in world states through the existence of world state propositions of the form intends($a$, $g_a$). These world state propositions record the fact that a character can have an intention, but do not indicate whether an intention is acted upon, nor do they capture which subsequent actions are executed in order for that character to act on the intention. Second, character intentions are recorded in frame of commitment data structures. Frames of commitment elaborate on the intention by also identifying which actions in the fabula plan the character is to perform in pursuit of the intention. Note that an interval of intentionality can contain more than one step with $g_a$ as an effect. This is necessary in the case where another action in the fabula plan undoes $g_a$ in the world and the condition must be reestablished.

IPOCL extends the definition of the POCL plan data structure to include frames of commitment. The definition of an IPOCL plan is as follows.

**Definition 4 (IPOCL Plan):** An IPOCL plan is a tuple $\langle S, B, O, L, C \rangle$ where $S$ is a set of plan steps, $B$ is a set of binding constraints on the free variables in the steps in $S$, $O$ is the set of ordering constraints on the steps in $S$, $L$ is a set of causal links between steps in $S$, and $C$ is a set of frames of commitment.

The sets $S$, $B$, $O$, and $L$ are defined in the standard way (e.g., Penberthy & Weld, 1992). The frames of commitment in $C$ are defined in Definition 3. See Figure 4 for an illustration of an IPOCL plan with a single frame of commitment and a motivating step for that frame. The actor of each step $s_i$ is indicated in parentheses. The IPOCL algorithm ensures that all story world characters that participate in a fabula plan appear to act believably with respect to intentionality. To satisfy this requirement, all character actions in an IPOCL plan (except those marked as not needing to be intentional) must be intentional in the final solution plan or happenings.

**Definition 5 (Action Intentionality):** An action in plan $P$ is intentional if it belongs to a frame of commitment in $P$.

Unintentional actions are not part of any interval of intentionality and are referred to as *orphans*. In order for an IPOCL plan to be considered complete, all actions – except for

happenings – must be part of at least one frame of commitment. A character action can belong to more than one interval of intentionality. The definition of plan completeness is as follows:

> **Definition 6 (Complete IPOCL Plan):** An IPOCL plan is complete if and only if (1) all preconditions of all plan steps are established, (2) all causal threats[6] are resolved, and (3) all plan steps that are not happenings are intentional.

Conditions 1 and 2 together make up the conventional definition of plan completeness, which can be termed causally complete. A fabula plan in IPOCL can be causally complete without being fully complete under Definition 6. When a plan is causally complete but not fully complete, then the plan contains orphans. If there are no ways to correct for the orphans, IPOCL backtracks to find another possible complete solution plan. Taken together, Definitions 4 and 6 directly address the high-level problem of finding a sound and believable sequence of actions that transforms an initial world state into a world state in which a goal situation holds.

## 4.3 Integrating Intentionality into Least-Commitment Planning

Frames of commitment are products of a process in which the planner tests the intentionality of character actions and revises the plan if necessary. IPOCL, as a refinement search process, uses an iterative, least-commitment process of identifying flaws in a plan and revising the plan to repair the flaws. This creates a tree-like search space in which leaf nodes are either complete plans (under Definition 6) or incomplete plans that cannot be repaired. Internal nodes are incomplete plans that have one or more flaws.

In addition to open conditions and causal threat flaws adopted from POCL we define three additional types of flaws:

> **Definition 7 (Open Motivation Flaw):** An open motivation flaw in plan $P$ is a tuple, $\langle c, p \rangle$, such that $c$ is a frame of commitment in $P$ and $p$ is the sentence $intends(a, g_a)$ such that $a$ is the character of $c$ and $g_a$ is the internal character goal of $c$.

> **Definition 8 (Intent Flaw):** An intent flaw in plan $P$ is a tuple $\langle s, c \rangle$ where $s$ is a step in $P$ and $c$ is a frame of commitment in $P$ such that $s \xrightarrow{p} s_j$ is a causal link in the plan, $s$ is not part of $c$, and $s_j$ is a step in $P$, is part of $c$, and the character of $s$ is the same as the character of $s_j$ and $c$.

> **Definition 9 (Intentional Threat Flaw):** An intentional threat flaw in plan $P$ is a tuple, $\langle c_k, c_i \rangle$, such that frame of commitment $c_k$ has an internal character goal that negates the internal character goal of another frame of commitment $c_i$.

Open motivation flaws reflect the fact that characters must appear motivated to have goals. That is, something must cause a character to commit to a goal. An open motivation flaw means that a plan has a frame of commitment whose interval of intentionality is not preceded by a motivating step. Intent flaws reflect the fact that a plan step, $s$, to be performed by a character can be part of a frame of commitment, $c$, held by that same character. That is, step $s$ causally establishes a precondition of some other step, $s_j$, which is part of $c$. The planner must non-deterministically decide whether the step is part of the frame of commitment. The next sections describe algorithms for identifying and repairing open motivation flaws and intent flaws. To facilitate this, the IPOCL algorithm, shown in Figure 5, is broken up into three parts: *causal planning*, *motivation planning*, and *intent planning*.

---

6. A causal threat occurs when, due to insufficient constraints of action ordering in a partially ordered plan, the effects of one action can potentially undo the preconditions of another action.

### 4.3.1 Causal Planning in IPOCL

The causal planning portion of the IPOCL algorithm implements the conventional POCL algorithm with the addition of a frame of commitment discovery phase. Causal planning occurs when there is an open condition that needs to be resolved. That is, some step $s_{need}$ has a precondition $p$ that is not satisfied by any causal link. The planner chooses a plan step $s_{add}$ whose effect $e$ can unify with $p$. This is accomplished by non-deterministically choosing an existing plan step or by instantiating a new action.

The frame of commitment discovery process is triggered by the changes in the plan (e.g. the addition of a causal link to the plan structure). If $s_{add}$ is a newly instantiated step, then there is the possibility that it is the final step (due to the backward-chaining nature of the planning algorithm) of some previously undiscovered character intention. If this is the case, then one of the effects of $s_{add}$, in addition to causally satisfying some open condition, is intended by the character specified to perform $s_{add}$. IPOCL non-deterministically chooses one of the effects of $s_{add}$ (or no effect, in the case where $s_{add}$ is not the final step of some yet-to-be-discovered intention). If an effect is chosen, then a new frame of commitment is constructed to record the character's commitment to achieving that effect in the world. Step $s_{add}$ is made to be the final step of the frame's interval of intentionality and a new open motivation flaw annotates the plan to indicate that the planner must find a motivating step.

Regardless of whether $s_{add}$ is newly instantiated or an existing plan step that is reused, the planner must consider the possibility that $s_{add}$ is part of an *existing* interval of intentionality. Steps can be performed as part of more than one intention; Pollack (1992) refers to this as *overloading*. IPOCL performs a search of the plan node for frames of commitment that $s_{add}$ can be part of. The search routine finds a set of frames $C''$ such that $c_j \in C''$ when one of the two following conditions holds:

1. The frame of commitment $c_j$ contains step $s_j$ such that $s_{add} \xrightarrow{p} s_j$ is a causal link in the plan and $s_{add}$ and $s_j$ are to be performed by the same character.

2. The frame of commitment $c_j$ contains step $s_j$ such that some frame $c_i \notin C''$ is in service of $s_j$ and $s_{add}$ is a motivating step for $c_i$. Frame $c_i$ is *in service of* step $s_j$ if the final step of $c_i$ has an effect that establishes a precondition of $s_j$, and $s_j$ is part of frame $c_k$, and $c_k \neq c_i$.

For each frame of commitment $c_i \in C''$, the plan is annotated with an intent flaw $\langle s_{add}, c_j \rangle$. By resolving these flaws, the planner will determine whether step $s_{add}$ becomes part of an existing frame's interval of intentionality.

---

**IPOCL** $(\langle S, B, O, L, C \rangle, F, \Lambda)$

The first parameter is a plan, with steps $S$, variable bindings $B$, ordering constraints $O$, causal links $L$, and frames of commitment $C$. $F$ is a set of flaws (initially open conditions for each literal in the goal situation). $\Lambda$ is a set of action schemata. Output is a complete plan according to Definition 6 or $fail$.

I. **Termination.** If $O$ or $B$ are inconsistent, fail. If $F$ is empty and $\forall s \in S, \exists c \in C \mid s$ is part of $c$, return $\langle S, B, O, L, C \rangle$. Otherwise, if $F$ is empty, fail.

II. **Plan Refinement.** Non-deterministically do one of the following.

- **Causal planning**

    1. **Goal selection.** Select an open condition flaw $f = \langle s_{\text{need}}, p \rangle$ from $F$. Let $F' = F - \{f\}$.

    2. **Operator selection.** Let $s_{\text{add}}$ be a step that adds an effect $e$ that can be unified with $p$ (to create $s_{\text{add}}$, non-deterministically choose a step sold already in $S$ or instantiate an action schema in $\Lambda$). If no such step exists, backtrack. Otherwise, let $S' = S \cup \{s_{\text{add}}\}$, $O' = O \cup \{s_{\text{add}} < s_{\text{need}}\}$, $B' = B \cup B_{\text{new}}$ where $B_{\text{new}}$ are bindings (e.g., assignments of ground symbols to variables) needed to make $s_{\text{add}}$ add $e$, including the bindings of $s_{\text{add}}$ itself, and $L' = L \cup \{\langle s_{\text{add}}, e, p, s_{\text{need}} \rangle\}$. If $s_{\text{add}} \neq s_{\text{old}}$, add new open condition flaws to $F'$ for every precondition of $s_{\text{add}}$.

    3. **Frame discovery.** Let $C' = C$.

        a. If $s_{\text{add}} \neq s_{\text{old}}$, non-deterministically choose an effect $e$ of $s_{\text{add}}$ or $e = nil$. If $e \neq nil$, construct a new frame of commitment $c$ with internal character goal $e$ and the character of $s_{\text{add}}$, let $s_{\text{add}}$ be part of $c$, let $C' = C \cup \{c\}$, create a new open motivation flaw $f = \langle c \rangle$, and let $F' = F \cup \{f\}$.

        b. Let $C''$ be the set of existing frames of commitment that can be used to explain $s_{\text{add}}$. For all $d \in C''$, create an intent flaw $f = \langle s_{\text{add}}, d \rangle$ and let $F' = F \cup \{f\}$.

    4. **Threat resolution**

        – **Causal threat resolution.** Performed as in II.3 in the POCL algorithm (Figure 1)

        – **Intentional threat resolution.** For all $c_1 \in C'$ and $c_2 \in C'$, such that the character of $c_1$ is the same as the character of $c_2$, $e_1$ is the goal of $c_1$, and $e_2$ is the goal of $c_2$, if $e_1$ negates $e_2$, non-deterministically order $c_1$ before $c_2$ or vice versa and for all $s_1 \in c_1$ and all $s_2 \in c_2$, $O' = O' \cup \{s_1 < s_2\}$ or $O' = O' \cup \{s_2 < s_1\}$.

    5. **Recursive invocation.** Call IPOCL $(\langle S', B', O', L', C' \rangle, F', \Lambda)$.

- **Motivation planning**

    1. **Goal selection.** Select an open motivation flaw $f = \langle c \rangle$ from $F$. Let $p$ be the condition of $c$. Let $F' = F - \{f\}$.

    2. **Operator selection.** Same as causal planning above, except
    $\forall s_i \in c, O' = O' \cup \{s_{\text{add}} < s_i\}$.

    3. **Frame discovery.** Same as for causal planning, above.

    4. **Threat resolution.** Same as for causal planning, above.

    5. **Recursive invocation.** Call IPOCL $(\langle S', B', O', L', C' \rangle, F', \Lambda)$.

- **Intent planning**

    1. **Goal selection.** Select an intent flaw $f = \langle s, c \rangle$ from $F$. Let $F' = F - \{f\}$.

    2. **Frame selection.** Let $O' = O$. Non-deterministically choose to do one of the following.

        – Make $s$ part of $c$. Let $s_{\text{m}}$ be the motivating step of $c$. $O' = O' \cup \{s_{\text{m}} < s\}$. For all $c_i \in C$ such that $c_i$ is ordered with respect to $c$, then for all $s_i \in c_i, O' = O' \cup \{s_i < s\}$ or $O' = O' \cup \{s < s_i\}$. For each $s_{\text{pred}} \in S$ such that $\langle s_{\text{pred}}, p, q, s \rangle \in L$ and $s_{\text{pred}}$ and $s$ have the same character, create an intent flaw $f = \langle s_{\text{pred}}, c \rangle$ and let $F' = F' \cup \{f\}$.

        – Do not make $s$ part of $c$.

    3. **Recursive invocation.** Call IPOCL $(\langle S, B, O', L, C \rangle, F', \Lambda)$.

---

Figure 5: The IPOCL algorithm.

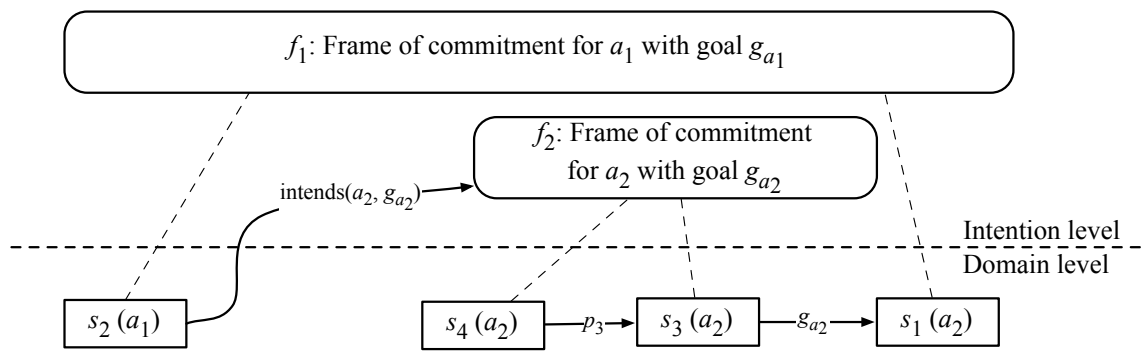

Figure 6: An IPOCL plan where one character is *contracted out* by another character.

Condition 1 indicates that if two actions, $s_j$ and $s_{add}$, are to be performed by the same character and the earlier action, $s_{add}$, establishes some condition in the world required for the later action, $s_j$, then a reasonable hypothesis is that both were part of the same intention. The intent flaw on the earlier action indicates that the planner must, at some point, decide whether to support this hypothesis by incorporating the actions into the same interval of intentionality or to reject the hypothesis by leaving the plan structure unchanged. Condition 2 indicates the situation where an agent requires a certain world state to be achieved to make its intentional actions feasible and this sub-goal is contracted out (e.g., Grosz & Kraus, 1996) to another agent. This occurs when a motivating action to be performed by one character causes another character to have an intention that is in service of the first character's actions, as illustrated in Figure 6. Character $a_1$ is to perform action $s_1$ in pursuit of goal $g_{a_1}$. Action $s_1$ has a single precondition that is satisfied by an action $s_3$ performed by character $a_2$ in pursuit of goal $g_{a_2}$. Action $s_2$ is the motivating action that causes character $a_2$ to have the goal to establish the precondition of step $s_1$. Since the motivating step is to be performed by character $a_1$, it is a candidate under Condition 2 to be incorporated into the frame of commitment of $a_1$.

Once frame discovery takes place, the planner must resolve any threats that were inadvertently introduced into the refined plan. There are two types of threats: causal threats and intentional threats. The standard POCL means of detecting and correcting causal threats is used (see Section 3). Intentional threats occur when a newly instantiated frame of commitment, $c_k$, has an internal character goal that negates the internal character goal of some other frame of commitment $c_i$ for the same character. Character actions in the fabula plan may be unordered with respect to one another and this allows for intervals of intentionality that are interleaved. While it is possible for an agent – or character – to hold conflicting desires, it is not rational for an agent to concurrently commit to conflicting desires (Bratman, 1987). In the case that a character has two frames with goals that negate each other, IPOCL corrects intentional threats by non-deterministically constraining the ordering of $c_i$ and $c_k$. The ordering of frames of commitment amounts to explicitly ordering the actions that are part of each frame to correspond to the ordering of $c_i$ and $c_k$. For more complicated cases in which the goals of $c_i$ and $c_k$ do not negate each other but in which plans causally interfere with each other, IPOCL relies on standard causal threat resolution to either order the action sequences of each plan while leaving the frames unordered, or to force the algorithm to backtrack. Some cases will not be identified or repaired without additional semantic and contextual reasoning.

### 4.3.2 MOTIVATION PLANNING IN IPOCL

The motivation planning portion of the IPOCL algorithm is responsible for ensuring that characters in the story world are motivated. A motivating step is a plan step in which one of its effects causes a character to commit to a goal. Repairing an open motivation flaw consists of non-deterministically finding a plan step with effect `intends(a, `$g_a$`)` – either by choosing an existing plan step or by instantiating an action schema and explicitly ordering that step before the plan steps that are part of the frame of commitment. Motivation planning is similar to causal planning except instead of establishing a causal link between two plan steps, it establishes a motivation link between a motivating step and a frame of commitment. Additionally, the motivating step for a frame of commitment is explicitly ordered before all other steps in the frames interval of intentionality. In the work presented here, a character agent cannot begin pursuing a character goal before it has committed to the goal. Motivation planning involves frame discovery and threat resolution phases that are identical to causal planning.

### 4.3.3 INTENT PLANNING IN IPOCL

The intent planning portion of the IPOCL algorithm determines interval membership for all character actions except those that are final steps for their intervals of intentionality. Intent planning repairs intent flaws. An intent flaw is a decision point that asks whether a plan step $s$ should be made part of the interval of some frame of commitment $c$. Unlike other flaws that are repaired by refining the structure of the plan, intent flaws are resolved by non-deterministically choosing one of the following:

- Make step $s$ part of the interval of $c$ and refine the plan structure to reflect the association.

- Do not make step $s$ part of the interval of $c$, remove the flaw annotation, and leave the plan structure unchanged.[7]

When the former is chosen, step $s$ becomes part of the interval of intentionality of frame $c$. When this choice is made, the interval of frame $c$ is updated appropriately and $s$ is explicitly ordered after the motivating step of frame $c$. Furthermore, the change in the step's membership status can have an effect on the membership of plan steps that precede $s$. Let $s_{pred}$ be an establishing step of $s$ – a step that precedes $s$ and is causally linked to $s$. The inclusion of step $s$ in the interval of frame $c$ also makes it possible for establishing steps to be included in the interval of $c$ if the following conditions hold:

- Step $s_{pred}$ is to be performed by the same character as $s$.

---

7. Because an intent flaw can be addressed by not making any revisions to the plan structure, an intent flaw is not strictly a flaw in the conventional sense. However, for the purpose of maintaining consistency with existing revision mechanisms, we find it useful to treat an intent flaw as a flaw up until the point that it is repaired.

- Step $s_{pred}$ is not a part of the interval of intentionality of $c$.

- The intent flaw, $f = \langle s_{pred}, c \rangle$ has not already been proposed and/or resolved.[8]

Intent flaws are created for each establishing step for which all three conditions hold. Intent planning thus operates in a spreading activation fashion. When one step becomes a member of a frame of commitment, an entire sequence of establishing steps may follow. This approach is necessary since frames of commitment can be created at any time during plan refinement. Intent flaws are not standard flaws in the sense that they mark a potential flaw instead of an actual flaw. We cannot determine at the time an action is instantiated whether it is necessary for that action to be part of an interval of intentionality. The frame of commitment it should belong to may not have been discovered yet, or it may not yet have been discovered that the action *can* be part of a frame of commitment due to adjacency requirements.

The propagation of intent flaws makes it possible for plan steps to become members of more than one frame of commitment, which is a desirable property of the IPOCL algorithm. Every time a character action – belonging to one frame of commitment – is used to satisfy an open condition of a successor action that belongs to a different frame of commitment, the system must non-deterministically decide whether the establishing action belongs to both frames of commitment or remains only a member of its original frame. The decision about interval membership also constrains the possible ordering of motivating steps for the frames of commitment involved because motivating steps are temporally ordered before all actions in the frame of commitment that the motivating step establishes. When a step becomes a member of more than one frame of commitment, the possible placement of motivating steps is constrained as in Figure 7 because the motivating step must occur before the earliest step in a frame of commitment.

One thing we have not yet discussed is how to handle *orphans*. An orphan is a step in the plan that does not belong to any interval of intentionality. Orphans are surreptitiously repaired when they are adopted into intervals of intentionality. This can happen when they causally establish conditions necessary for other, intentional actions. Orphaned actions cannot be repaired directly because frames of commitment are discovered opportunistically instead of instantiated in a least-commitment approach (as plan steps are). If there are orphans remaining that have not been surreptitiously repaired by the time the planning process completes, then the planner must backtrack.

## 4.4 An Example

The IPOCL algorithm is illustrated by the following story about an arch-villain who bribes the President of the United States with a large sum of money. The example traces a single path through the fabula plan search space generated by IPOCL. The initial plan node contains only the initial state step and goal situation step. The initial state contains propositions describing the state of the world before the story begins. The goal situation contains a single proposition, `corrupt(President)`, which describes what must be different about the world after the story is complete. The story that will be generated by IPOCL is, in effect, the story about how the President becomes corrupt.

---

8. The inclusion of this condition ensures the *systematicity* of the algorithm since there can be more than one causal link between $s_{pred}$ and $s$. A search algorithm is systematic if it is guaranteed to never duplicate a portion of the search space.

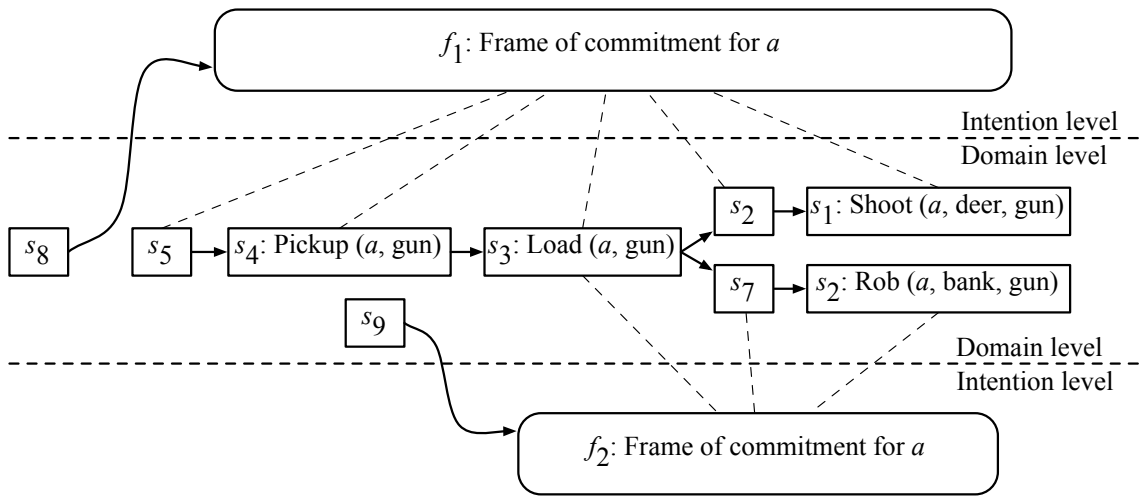

Figure 7: An IPOCL plan with overlapping intervals of intentionality for a single character.

The goal proposition `corrupt(President)` is non-deterministically established by instantiating a new character action, `Bribe(Villain, President, $)`, which states that the Villain character will bribe the President character with some money. The `Bribe` action was chosen because it has `corrupt(President)` as an effect. From the planners perspective, the `Bribe` action is causally motivated by the open condition of the goal situation. Upon instantiation of the `Bribe` action, frame discovery is invoked. The effects of the `Bribe` action are:

- `corrupt(President)` – the President is corrupt.

- `controls(Villain, President)` – the Villain exerts control over the President.

- `has(President, $)` – the President has the money.

- ¬`has(Villain, $)` – the Villain does not have the money.

From the audience's perspective, any of these effects can be a reason why the Villain performs the actions in the story.

The planner non-deterministically chooses `controls(Villain, President)` as the character goal for the Villain character. Note that in this case the goal of the Villain differs from the outcome of the story although the same action satisfies both conditions. There is no reason why the planner could not have chosen `corrupt(President)` as the character goal for the Villain. It is assumed here that either the plan cannot be completed if the alternative is chosen or that some heuristic function has evaluated all options and determined that villains are more likely to want control over the President than anything else. Given the choice made, the planner constructs a frame of commitment for the Villain character and makes the `Bribe` action the final step in the frame's interval of intentionality. Even with the new frame of commitment, the plan is still flawed since there is no reason for the Villain character to have the goal of controlling the President. That is, the Villain needs to *form* the intention to appear believable to the audience. An open motivation flaw indicates that some action in the plan must satisfy the condition `intends(Villain, controls(Villain, President))` on the frame of commitment.

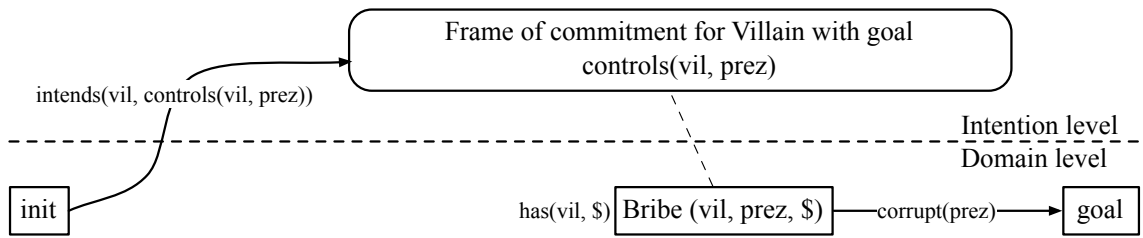

Figure 8: Example narrative plan after discovering the one action and corresponding frame of commitment.

Since there are no other frames of commitment for the Villain, no intent flaws occur. The `Bribe` action, however, has a precondition `has(Villain, $)` that becomes an open condition; the Villain character must have the money if he is to bribe the President with it. The planner chooses to repair the open motivation flaw on the single frame of commitment first and non-deterministically chooses the initial state to satisfy the open motivation condition. This illustrates a situation where the intention of a character in the story world is encoded as part of the initial conditions. While it does not have to be this way, the domain engineer that specified the inputs to IPOCL has decided that no further motivation for the Villain to want to control the President is needed. While this may not be the most satisfactory solution, it is a valid solution. The partial plan at this point is shown in Figure 8.

The open condition `has(Villain, $)` on the `Bribe` action is considered next. To repair this flaw, the planner non-deterministically instantiates a new character action `Give(Hero, Villain, $)` in which the Hero character gives the Villain the money. The planner must consider, from the audience's, perspective, why the Hero character gives the money to the Villain. The planner inspects the effects of the `Give` action:

- `has(Villain, $)` – the Villain has the money.

- ¬`has(Hero, $)` – the Hero does not have the money.

The planner non-deterministically chooses `has(Villain, $)` as the goal that the Hero is attempting to achieve. A new frame of commitment for the Hero's goal is created. Note that the Hero's intention matches the open condition that the Give action was instantiated to satisfy. This indicates that the Hero's commitment is in service to the `Bribe` action.

An open motivation flaw is created that corresponds to the new frame of commitment. There are many actions that will establish the Hero's intention that the Villain has the money: the Villain might persuade the Hero if they are friends, or the Villain might coerce the Hero. The latter, `Coerce(Villain, Hero, has(Villain, $))`, is chosen by the planner: the Villain character coerces the Hero character into having the goal `has(Villain, $)`.

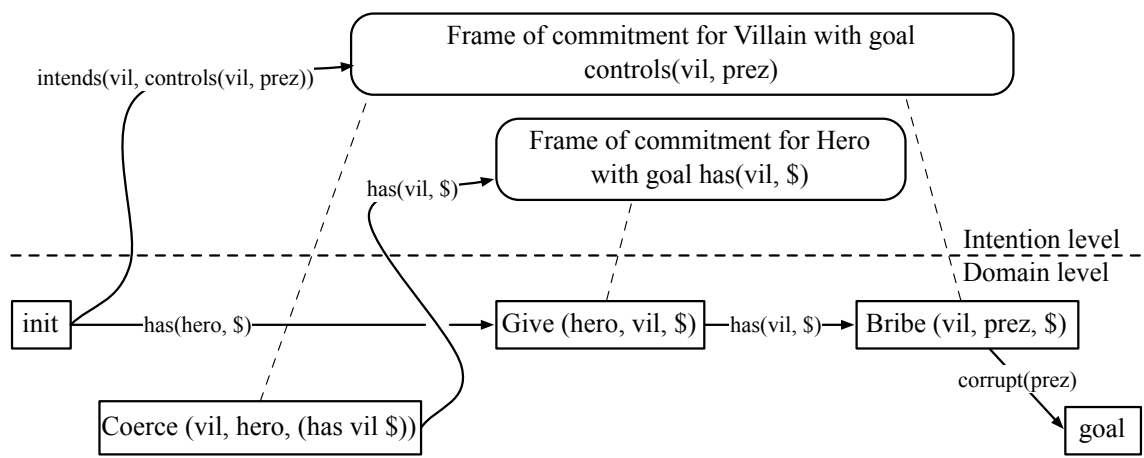

Figure 9: Solution IPOCL plan graph for the example narrative.

At this point, the planner must determine why the Villain coerces the Hero. There are several possibilities. First, frame discovery comes into play to determine if the Villain intends any of the effects of the `Coerce` action. Assume the only effect of the `Coerce` action is `intends(Hero, has(Villain, $))`. The planner can select this effect and construct a new frame of commitment specifying that the Villain intends that the Hero intends that the Villain has the money. Another option is to leave the `Coerce` action an orphan for the time being. Let us suppose that this is the course that the planner chooses. A search of the current plan structure indicates that the `Coerce` action can be part of the Villain's existing commitment to control the President. This is possible because `Coerce` is a motivating step for the Hero's frame of commitment and the Heros frame of commitment is in service to the `Bribe` action, which is part of the Villain's frame of commitment. An intent flaw associating the `Coerce` action with the Villain's existing frame of commitment is created. Eventually, the spreading activation of *intent planning* will associate `Coerce` with the Villain's frame of commitment. The plan structure at this point is shown in Figure 9. Any remaining flaws are handled by conventional causal planning.

## 4.5 Complexity of the IPOCL Algorithm

The computational complexity of the IPOCL algorithm is $\mathcal{O}(c(b(e+1)^a)^n)$, where

- $n$ is the depth of the search space,

- $b$ is the number of ways that an action can be instantiated (e.g., the number of permutations of legal parameter bindings),

- $e$ is the number of effects of an instantiated action, and

- $a$ is the number of actors in an instantiated action.

The worst-case branching factor of the IPOCL search space is $b(e + 1)^a$. The factor, $(e + 1)$ signifies that if a new frame of commitment is being constructed, the planner must choose between the $e$ effects of the action (plus one to signify the condition where no effect is chosen). The exponent $a$ reflects the fact that if multiple characters are intentionally participating in an action, then each of those characters can have distinct intentions for performing that action. For example, the action `Marry(?groom, ?bride, ?place)` has six effects (see the Appendix for the action schema) and two intentional actors (`?groom` and `?bride`).

The depth of the IPOCL search space $n$ is the number of open condition flaws, open motivation flaws, intent flaws, causal threats, and intentional threats that are repaired. In the worst-case, for every newly instantiated step in the plan IPOCL also creates a new frame of commitment and a corresponding open motivation flaw. If $n_{\text{POCL}}$ is the depth of a solution in the search space of POCL planning problem and $n_{\text{IPOCL}}$ is the depth of the corresponding solution in the search space on an IPOCL fabula planning problem, then $n_{\text{IPOCL}}$ is bounded by the function $n_{\text{IPOCL}} = 2n_{\text{POCL}}$.

A narrative was generated for evaluation purposes (see Section 5). The narrative, rendered into natural language, is given in Figure 13 and the plan data structure is presented graphically in the Appendix (Figure 15). This complete fabula plan exists at a depth of $n = 82$. The average branching factor for this domain is $\sim 6.56$, with the worst branching factor for any given node being 98. The node with 98 children is the first flaw that the planner solves for: `married(Jafar, Jasmine)`, which is solved by instantiating the action `Marry(Jafar, Jasmine, ?place)`. The operator schema has six effects. The two characters are both intentional actors meaning there can be up to two frames of commitment generated. Finally, the parameter `?place` can be bound in two ways. Note that our implementation of IPOCL uses constraint propositions to generate a child node for each legal permutation of parameter bindings. When we provide a domain-specific heuristic evaluation function that favors plan structures with certain preferred character goals (for example, "Jafar intends that Jasmine is dead" is not included), then the complete plan is generated in approximately 12.3 hours (approximately 11.6 hours were spent in garbage collection) on an Intel Core2 Duo 3GHz system with 3GB of RAM and 100GB of virtual memory running Allegro CL®8.0. Under these conditions, IPOCL generates 1,857,373 nodes and visits 673,079 nodes. When the algorithm is run with only a domain-independent heuristic adopted from classical planning (e.g., number of flaws plus plan length), the problem cannot be solved before the system runs out of virtual memory. The Appendix gives details on the domain, fabula planning problem, and heuristic used.

Practical experience with the IPOCL algorithm suggests that better heuristic evaluation functions are needed to guide the search process. Without sufficient heuristic functions, the behavior of IPOCL devolves to nearly breadth-first. Practical experience with the algorithm also suggests that it is difficult to write heuristic functions that practically distinguish between sibling nodes. The problem of defining heuristic functions that distinguish between sibling nodes in the plan search space arises in all POCL algorithms, but is exacerbated in IPOCL due to the increased number of structural features that need to be distinguished.

## 4.6 Limitations and Future Work

As an algorithm that solves the fabula generation problem, IPOCL has been demonstrated to generate sound narrative structures that support believability through the enforcement of character intentions. IPOCL is able to achieve this by effectively decoupling the concept of character intentions from those of author intentions. Consequently, intentionality of character actions must be opportunistically discovered at the time that actions are discovered. The opportunistic discovery of character intentions during action instantiation significantly increases the branching factor to the detriment of the ability to generate long narratives. However, we feel that opportunistic discovery of intentions is a vital part of expanding the space of narratives that can be searched to include those that have logical causal progression and also have well-motivated and thus more believable characters. An alternative is to use a grammar (cf., Rumelhart, 1975), hierarchical task networks (cf., Sacerdoti, 1977), or other form of hierarchical decomposition (cf., Young, Pollack, & Moore, 1994) such that intentions are dealt with at one level of abstraction and specific character actions dealt with at the primitive level. However, using grammars, HTNs, or other decompositional techniques to generate narrative requires reasoning at higher levels of abstraction than the action and introduces potentially rigid top-down structuring of plot that can limit the system's ability to find solutions that might exist but cannot be described by the grammar/task-network.

There are additional limitations that need to be addressed. First, while the IPOCL algorithm asserts that all non-happening character actions must be part of a frame of commitment, and therefore motivated by an event (or the initial state), IPOCL also assumes that each frame of commitment's interval of intentionality terminates in an action that successfully achieves the goal of the frame of commitment. Essentially, every character acts according to an intention and every intention is achieved. This inherently limits the types of narratives that can generated. For example, narratives in which a character tries to achieve a goal but fails several times before finally succeeding are unlikely. Narratives in which one character – a hero – defeats another – a villain – cannot be generated. Although, it is possible to generate a narrative in which the villain first achieves his goal and then the hero achieves his goal (thus defeating the villain).

The inability to consider actions that support intentions that are never achieved appears to be an inherent limitation of our partial-order planning approach. In particular, the backward-chaining nature of the algorithm biases the approach toward *explaining* actions. To ensure soundness, causal threats are eliminated or backtracking occurs. It is possible that a forward-chaining approach could resolve this issue, but only at the expense of promiscuous intention generation. One way to force the algorithm to consider narrative structures in which one character defeats another or in which a character fails several times before succeeding is to seed the plan space with intermediate author goals indicating sets of states through which all solutions must pass (Riedl, 2009). This approach, however, presupposes that the human author knows, wants, or can predict some of the resultant narrative structure.

As mentioned in Section 4.3.1, IPOCL currently only has weak mechanisms to detect or prevent contradictory intentions for a character. Better heuristics may help control for the situations that are not resolved through ordering of actions or ordering of frames of commitment. It is possible to extend the algorithm to include common-sense reasoning or semantic analysis at the frame of commitment level. However, this work has not been done.

In general, better heuristics are needed. Heuristics can be divided into *domain-dependent* and *domain-independent* heuristics. Domain-dependent heuristics, in this case, refer to those that employ knowledge about the characters, setting, or preferences over the narrative structure. For example, to generate the example in Figure 13, we use a heuristic that penalizes narratives that contain character goals that we thought unreasonable based on our intuitions about characters and the types of stories that we sought. Domain-independent heuristics are more difficult to identify but might include preferences for fewer frames of commitment with longer action sequences. Domain-independent heuristics that can reward narrative structures with dramatic arc will likely require complex models of narrative psychology such as those described by Gerrig and colleagues (Gerrig, 1993; Gerrig & Bernardo, 1994) and implemented by Fitzgerald, Kahlon, and Riedl (2009) and may not work on intermediate, incomplete narratives.

## 5. An Evaluation of Character Intentionality in IPOCL-Generated Fabula Plans

In order to perform an empirical evaluation of a reader's perception of character intentionality in IPOCL-generated fabulas, we designed an objective evaluation procedure based on question-answering in order to reveal a readers understanding of character intentions without the use of subjective questionnaires (Riedl & Young, 2005). The goal of the evaluation was to determine if IPOCL-generated fabulas supported the cognitive processes that readers apply to comprehend character actions better than fabulas generated by conventional planning algorithms. The evaluation procedure is outlined as follows. Two planning-based algorithms were used to generate plans to be interpreted as fabulas: the IPOCL algorithm and a conventional POCL planning algorithm. Each planner was provided identical initialization parameters. The first plan generated by each algorithm was selected to be presented to study participants. Because the plans must be read, a simple natural language generation process was used to produce natural language text from each fabula plan. Recall that the purpose of a fabula plan is not to be executed by a plan execution system, but to contain temporal event information to be told as a story. Participants were recruited and randomly assigned to one of two groups. Participants in the *POCL* group read the POCL-generated narrative text. Participants in the *IPOCL* group read the IPOCL-generated narrative text. A variation of the question-answering protocol from the work of Graesser et al. (1991) was used to elicit participants' mental models of the narratives. In particular, we focused on "why" questions that elicit understanding of story world character goals and motivations.

How do we evaluate question-answering performance across groups? QUEST (Graesser et al., 1991) takes a graphical representation of a story and reliably predicts the question-answering performance of a human who might also read the story. One of the implicit assumptions behind QUEST is that it has been provided a well-structured story that has contained within the story's narrative structure all the answers to any question one might ask about character goals and motivations. We exploit that assumption as a means of measuring how well a story actually supports a human reader's reasoning about character goals and motivations. That is, if a story does not support human comprehension, we should see

> Once there was a Czar who had three lovely daughters. One day the three daughters went walking in the woods. They were enjoying themselves so much that they forgot the time and stayed too long. A dragon kidnapped the three daughters. As they were being dragged off, they cried for help. Three heroes heard the cries and set off to rescue the daughters. The heroes came and fought the dragon and rescued the maidens. Then the heroes returned the daughters to their palace. When the Czar heard of the rescue, he rewarded the heroes.

Figure 10: An example story from the work of Graesser et al. (1991).

this manifested in human question-answering performance. QUEST knowledge structures can be translated into fabula plans and vice versa (Christian & Young, 2004). From QUEST knowledge structures automatically generated from fabula plans, we run QUEST to predict question-answering performance and compare QUEST predictions to actual performance. We expect to see that IPOCL-generated narratives are more understandable than the alternative; we should find a greater correspondence between QUEST and actual performance in the *IPOCL* condition than we find in the *POCL* condition.

## 5.1 The QUEST Model of Question-Answering

The QUEST model (Graesser et al., 1991) accounts for the goodness-of-answer (GOA) judgments for questions asked about passages of prose. One application of the QUEST model is to show that people build cognitive representations of stories they read that capture certain relationships between events in a story and the perceived goals of the characters in the story (Graesser et al., 1991). QUEST knowledge structures can be represented visually as directed graphs with nodes referring to either story events (typically character actions) or character goals. Directed links capture the relationship between story event nodes in terms of causality and the relationship between events and character goals in terms of intentionality. A reader's cognitive representation of the story is queried when the reader answers questions about the story. The types of questions supported by the QUEST model are: why, how, when, enablement, and consequence. For example, the story in Figure 10 (Graesser et al., 1991, Fig. 1) has the corresponding QUEST knowledge structure shown in Figure 11 (Graesser et al., 1991, Fig. 2). There are two types of nodes in the QUEST knowledge structure: event nodes, which correspond to occurrences in the story world, and goal nodes, which correspond to goals that characters have. The links between nodes capture the different types of relationships between events and character goals.

- Consequence (C): The terminal event node is a consequence of the initiating event node.

- Reason (R): The initiating goal node is the reason for the terminal goal node.

- Initiate (I): The initiating event node initiates the terminal goal node.

- Outcome (O): The terminal event node is the outcome of the initiating goal node.

- Implies (Im): The initiating event node implies the terminal event node.

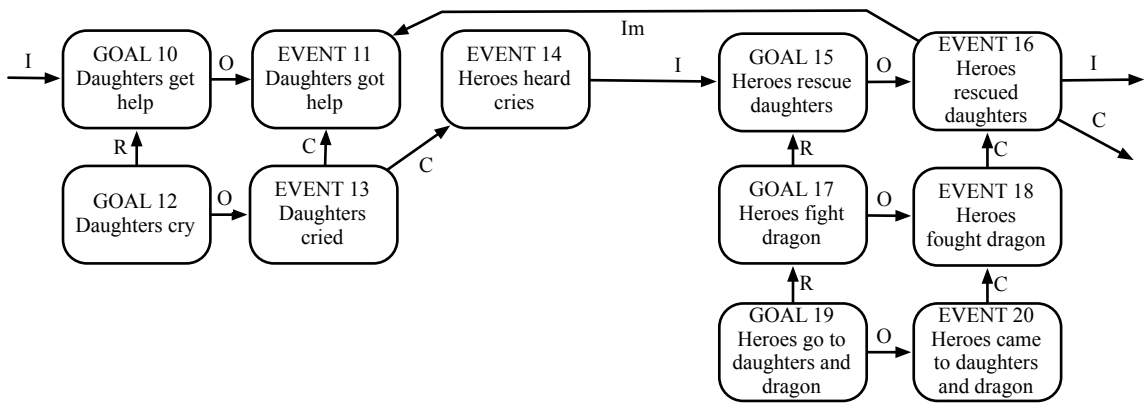

Figure 11: An example of a QUEST model of the Czar and the Daughters story from Graesser et al. (1991).

The QUEST model defines arc search procedures for each type of question (e.g. why, how, when, enablement, and consequence). The arc search procedures, starting at the queried node, distinguish between legal answer nodes and illegal answer nodes. That is, only nodes reachable by the arc search procedures are legal answer nodes. The legality of answers and the weight of structural distance correspond to GOA judgments of human story readers.

## 5.2 Procedure

The procedure involves comparing subject question-answering performance to QUEST question-answering predictions in two conditions. The *POCL* condition is based on narrative structures generated by a conventional POCL planning algorithm. The *IPOCL* condition is based on narrative structures generated by the IPOCL algorithm. Both planners were initialized with identical information defining a story world. The story world was based loosely on the story of *Aladdin*. The initial parameters included the following:

- An initial state that defines the story world, including (a) locations, (b) objects, (c) characters, and (d) the relevant initial relationships between all of the above. Story world characters include Aladdin, King Jafar, Princess Jasmine, a dragon, and a genie.

- A library of operators defining the events that can be performed by story world characters.

- An outcome: Jasmine and Jafar are married and the genie is dead.

Note that even though the initialization parameters were identical in both conditions, there are differences in how the respective planners handle the parameters. In particular, IPOCL makes use of additional information in action schemata about actors and whether an action can be a happening. The POCL planner ignored this information, which did not impact its ability to find a valid solution. The initialization information used by IPOCL but ignored by the POCL planner are as follows. First, all operators must specify which of the parameters are the intentional actors (and not characters being acted upon). Second, some operators are tagged as happenings. Finally, some operators have effects of the form `intends(a, p)` where $a$ is a variable that can be bound to a ground symbol representing a character, and $p$ is a variable that can be bound to a literal that becomes one of the character's internal goals. Effects of this form are used by the IPOCL implementation during motivation planning to ensure there are actions that cause frames of commitment. Since no operators have preconditions of this form, the POCL planner does not utilize this information. The Appendix lists the entire set of initialization parameters used in the evaluation.

The fabula plans generated by the two planning algorithms are shown in the Appendix. For the plans to be human-readable, each plan was input into the Longbow discourse planner (Young et al., 1994). Longbow results in a plan consisting of communicative acts such as `describe-character` and `describe-event` for conveying the temporally ordered information of the narrative. The discourse plan steps were then rendered into natural language using a simple template-matching procedure. The resulting narrative texts for the *POCL* condition and *IPOCL* condition are shown in Figures 12 and 13, respectively. Similarities between the two narratives make a comparison study possible. Specifically, the set of events in the IPOCL-generated narrative are a superset of the events in the POCL-generated narrative. There is one distinct action ordering difference between the two fabula plans: in the *IPOCL* condition only the event where the King falls in love with Jasmine is temporally constrained to occur first, but in the *POCL* condition the ordering of this event is under-constrained and falls late in the text. In the *POCL* condition, had this event come earlier, some participants may have inferred a relationship between the king falling in love and Aladdin's actions even though there is no actual relationship in the generated QUEST graph. We believe that the ordering had an insignificant impact on the results.

The fabula plans were also converted to structures that QUEST can use to predict question-answering performance. We use a procedure described by Christian and Young (2004). Their algorithm for generating QUEST graph structures from a plan has been only evaluated for POCL plans involving a single character.[9] IPOCL plans, however, contain additional structures such as frames of commitment and motivation links that are not part of conventional plan representations. Consequently, the algorithm for generating a QUEST graph structure from a plan was extended to take into consideration IPOCL plans. An additional study by the authors (not reported) determined that QUEST knowledge structures derived from IPOCL plans with the extended algorithm significantly predict question-answering judgments when structural distance is ignored ($p < 0.0005$). The modifications to Christian and Young's (2004) algorithm are beyond the scope of this paper, but details can be found in the work of Riedl (2004).

The evaluation involved a questionnaire in which participants read a story and then make goodness-of-answer (GOA) judgments about pairs of question and answers. A question-answer pair has a "why" question about an intentional action performed by a character in the story and a possible answer. For example, the question, "Why did Aladdin slay the

---

9. Christian and Young (2004) compare DPOCL plans to QUEST knowledge structures. DPOCL is a decompositional, partial order causal link planning algorithm (Young et al., 1994) that extends the conventional POCL algorithm by explicitly representing hierarchical relationships between abstract and primitive planning operators.

There is a woman named Jasmine. There is a king named Jafar. This is a story about how King Jafar becomes married to Jasmine. There is a magic genie. This is also a story about how the genie dies.

There is a magic lamp. There is a dragon. The dragon has the magic lamp. The genie is confined within the magic lamp. There is a brave knight named Aladdin. Aladdin travels from the castle to the mountains. Aladdin slays the dragon. The dragon is dead. Aladdin takes the magic lamp from the dead body of the dragon. Aladdin travels from the mountains to the castle. Aladdin hands the magic lamp to King Jafar. The genie is in the magic lamp. King Jafar rubs the magic lamp and summons the genie out of it. The genie is not confined within the magic lamp. The genie casts a spell on Jasmine making her fall in love with King Jafar. Jasmine is madly in love with King Jafar. Aladdin slays the genie. King Jafar is not married. Jasmine is very beautiful. King Jafar sees Jasmine and instantly falls in love with her. King Jafar and Jasmine wed in an extravagant ceremony.

The genie is dead. King Jafar and Jasmine are married. The end.

Figure 12: Text of story in control condition.

There is a woman named Jasmine. There is a king named Jafar. This is a story about how King Jafar becomes married to Jasmine. There is a magic genie. This is also a story about how the genie dies.

There is a magic lamp. There is a dragon. The dragon has the magic lamp. The genie is confined within the magic lamp.

King Jafar is not married. Jasmine is very beautiful. King Jafar sees Jasmine and instantly falls in love with her. King Jafar wants to marry Jasmine. There is a brave knight named Aladdin. Aladdin is loyal to the death to King Jafar. King Jafar orders Aladdin to get the magic lamp for him. Aladdin wants King Jafar to have the magic lamp. Aladdin travels from the castle to the mountains. Aladdin slays the dragon. The dragon is dead. Aladdin takes the magic lamp from the dead body of the dragon. Aladdin travels from the mountains to the castle. Aladdin hands the magic lamp to King Jafar. The genie is in the magic lamp. King Jafar rubs the magic lamp and summons the genie out of it. The genie is not confined within the magic lamp. King Jafar controls the genie with the magic lamp. King Jafar uses the magic lamp to command the genie to make Jasmine love him. The genie wants Jasmine to be in love with King Jafar. The genie casts a spell on Jasmine making her fall in love with King Jafar. Jasmine is madly in love with King Jafar. Jasmine wants to marry King Jafar. The genie has a frightening appearance. The genie appears threatening to Aladdin. Aladdin wants the genie to die. Aladdin slays the genie. King Jafar and Jasmine wed in an extravagant ceremony.

The genie is dead. King Jafar and Jasmine are married. The end.

Figure 13: Text of story in experimental condition.

dragon?" might be paired with the answer, "Because King Jafar ordered Aladdin to get the magic lamp for him." The participants were asked to rate the goodness of the answer for the given question on a four-point scale ranging from "Very bad answer" to "Very good answer." The participants were shown examples of a question-answer pairs before the rating task began, but were not otherwise given a definition of "good" or "poor" or trained to make the judgment. Participants rated the GOA of a question-answer pair for every combination of goal nodes in the QUEST knowledge structure for the story. The *POCL* condition questionnaire had 52 question-answer pairs while the *IPOCL* condition questionnaire had 82 question-answer pairs due to the increased story plan length. Participants were asked to read the story text completely at least once before proceeding to the ratings task and were allowed to refer back to the original text at any time during the rating task.

For each narrative, QUEST was used to predict whether question-answer pairs would be considered as "good" or "poor" based on the arc search procedure following forward reason arcs, backward initiate arcs, and backward outcome arcs (Graesser et al., 1991). The hypotheses of the experiment were as follows.

**Hypothesis 1** Participants in the *IPOCL* condition will have higher mean GOA judgment ratings for question-answer pairs identified by QUEST as being "good" than participants in the *POCL* condition.

**Hypothesis 2** Participants in the *IPOCL* condition will have lower mean GOA judgment ratings for question-answer pairs identified by QUEST as being "poor" than participants in the *POCL* condition.

If the actual question-answering performance of participants results in statistically higher GOA ratings for question-answer pairs judged by QUEST to be "good," then there is greater correspondence between QUEST predictions and actual performance. Likewise, if actual question-answering performance of participants results in statistically lower GOA ratings for question-answer pairs judged by QUEST to be "poor," then there is greater correspondence between QUEST predictions and actual performance. Poor correspondence between QUEST predictions and actual question-answering performance is an indication that a narrative lacks structure that supports human understanding of character goals, intentions, and motivations.

Thirty-two undergraduate students in the Computer Science program at North Carolina State University participated in the study. All participants were enrolled in the course *Game Design and Development* and were compensated for their time with five extra credit points on their final grade in the course.

## 5.3 Results and Discussion

Each question-answer pair in each questionnaire was assigned a "good" rating or a "poor" rating based on the QUEST prediction. "Good" question-answer pairs were assigned a value of 4 and "poor" question-answer pairs were assigned a value of 1. Human GOA ratings of question-answer pairs were also assigned values from 1 to 4 with 1 corresponding to "Very poor answer" and 4 corresponding to "Very good answer." The results of participants' answers to questionnaire answers are compiled into Table 1. The numbers are the mean GOA ratings for each category and each condition. The numbers in parentheses are standard deviations for the results. Mean human question-answering performance is more in agreement with QUEST when the mean GOA ratings for question-answer pairs categorized as "good" is closer to 4 and the mean GOA ratings for question-answer pairs categorized as "poor" is closer to 1.

A standard one-tailed *t*-test was used to compare the mean GOA rating of "good" question-answer pairs in the *IPOCL* condition to the mean GOA rating of "good" question-answer pairs in the *POCL* condition. The result of the *t*-test with 15 degrees of freedom yields $t = 1.6827$ ($p < 0.0585$). This result is strongly suggestive that Hypothesis 1 is supported.

| Condition | Mean GOA rating for "good" question-answer pairs (standard deviation) | Mean GOA rating for "poor" question-answer pairs (standard deviation) |
|---|---|---|
| *IPOCL* | 3.1976 (0.1741) | 1.1898 (0.1406) |
| *POCL* | 2.9912 (0.4587) | 1.2969 (0.1802) |

Table 1: Results of the evaluation study.

A standard one-tailed *t*-test was used to compare the mean GOA rating of "poor" question-answer pairs in the *IPOCL* condition to the mean GOA rating of "poor" question-answer pairs in the *POCL* condition. The result of the *t*-test with 15 degrees of freedom yields $t = 1.8743$ ($p < 0.05$). Participants in the *IPOCL* condition had significantly lower GOA ratings for "poor" question-answer pairs than participants in the *POCL* condition. Hypothesis 2 is supported.

It is interesting to note that the standard deviation for results in the *POCL* condition for "good" question-answer pairs was high. Further analysis reveals that human participants are likely to judge a question-answer pair as "good" if there is lack of evidence against the possibility that the character action might have been intentional. We speculate that reader/viewers simultaneously consider multiple hypotheses explaining character behavior until they are disproved. Regardless of the content of any communicative act, one will always be able to provide a more or less plausible explanation of the meaning (Sadock, 1990).

There were a couple of limitations to note. We did not control for narrative length. It is possible that the effects we measured were a result of narrative length instead of improved narrative structure generated by IPOCL. We believe this to be unlikely, but future evaluations should add "filler sentences" to the *POCL* condition narrative that do not impact character intentionality so that the control narrative matches the length of the *IPOCL* condition. According to the narrative comprehension theories of Graesser et al. (1994) and Trabasso and colleagues (Trabasso & Sperry, 1985; Trabasso & van den Broek, 1985), such filler sentences will not be included in the reader's mental model in a meaningful way because they will not be causally related to other concepts in the mental model of the narrative. Consequently, we felt that it was safe to leave the filler sentences out. Another limitation is that the model of discourse used in the Longbow discourse planner was simplistic. Specifically, explicit statements about character intentions were incorporated into the narrative text in the experimental condition due to an overly promiscuous discourse model used by the Longbow discourse planner. We believe that our results would be the same if these explicit statements were excluded because human readers are very good at inferring intentions from stories (Graesser et al., 1991, 1994). However, to be complete we would have to control for such artifacts from discourse generation.

We conclude that there is strong evidence that the narrative in the experimental condition supported reader comprehension of character goals, intentions, and motivations better than the narrative in the control condition. Since both were generated from identical initial-ization parameters, the most significant independent variable is the generation algorithm. We infer that the improvement of the *IPOCL* condition over the *POCL* condition is due to enhancements to the automated story generation capability introduced in the IPOCL algorithm.

## 6. Conclusions

The objective of the research presented here is to develop an approach to the generation of narrative fabula that has the properties of supporting audience perception of character intentionality and causal plot progression. An informal analysis of related work suggests that narrative generation systems can be categorized as using *simulation-based* or *deliberative* approaches. Simulation-based approaches tend to produce narratives with reasonable character believability but are unlikely to produce narrative with globally coherent plots. This is due to the fact that simulation-based approaches model characters and attempt to optimize character decisions in any given moment. Thus simulation-based approaches are prone to local maxima. Deliberative approaches reviewed in this article do not directly consider character intentions but are otherwise more likely to produce narratives with causally coherent plot progressions. This is due to the fact that deliberative narrative generation systems tend to reason about the entire plot instead of separately about characters. In our informal analysis, we did not see any evidence that a deliberative system cannot reliably produce narrative structures with character believability (especially character intentionality). We use a refinement search approach to construct the entire fabula from the perspective of the author. Our approach is consequently a deliberative one. We favor refinement search because partial-order plans appear to be a good representation for the fabula of a narrative.

In our analysis, algorithms that solve the planning problem are not sufficient for generating narratives with character intentionality because planning algorithms are conventionally designed to provide a singular agent with the ability to achieve a singular goal. Stories are more likely than not to involve multiple agents – characters – who are not necessarily cooperating to achieve a singular goal state. Accordingly, we developed a deliberative fabula generation algorithm that reasons about the understandability of characters from the perspective of a hypothetical audience. The IPOCL fabula generation algorithm treats any potential solution as flawed unless the audience is capable of understanding what individual (and potentially conflicting) goals each character has and what motivated the characters to adopt those goals throughout the progression of the narrative. IPOCL contains routines for repairing character intentionality flaws by non-deterministically attributing goals to characters and then generating event sequences that motivate those goals.

In adopting the approach to narrative generation in the IPOCL algorithm, we realize several limitations. First, IPOCL is incapable of producing narrative structures in which a character fails to achieve their goals. Character failure of this sort is a natural part of most stories and is especially important in comedy and tragedy (Charles et al., 2003). Unfortunately, planners do not produce plans that fail – e.g. cannot execute to completion – because the planner will prune that branch of the search space and backtrack. Conflict can arise between characters when characters adopt contradictory goals. However, each character will succeed in achieving their goal, although this will happen serially because conflicting frames of commitment will be temporally ordered. Second, in our work to date, we have assumed that fabula and sjuzet can be reasoned about distinctly. That is, once a fabula is generated indicating what a narrative is about, a separate process can reason about how the narrative should be told. This may suffice for simple telling of generated narratives. Intuitively, in order to achieve more sophisticated effects on an audience such as suspense, one might have to consider how a narrative can be told while the generator is determining what should be told.

We believe that the work reported here represents a step towards achieving greater capability in computer systems to generate fictional narratives for communication, entertainment, education, and training. It is an incremental step, building from established artificial intelligence technologies – planning – and cognitive science principles. Non-subjective empirical evidence suggests that we have achieved improvement in narrative generation over alternative, conventional planners. Furthermore, we believe we have created a framework on which we can continue to make incremental improvements to narrative generation capabilities. For example, we have been able to incorporate the ability to handle folk psychological models of character personality (Riedl & Young, 2006). A system that can generate stories is capable of adapting narrative to the user's preferences and abilities, has expanded replay value, and is capable of interacting with the user in ways that were not initially envisioned by system designers. Narrative generation is just one example of how instilling computational systems with the ability to reason about narrative can result in a system that is more capable of communicating, entertaining, educating, and training.

## Appendix A.

This appendix contains details about the Aladdin planning domain used for the evaluation, including planning problem specification, heuristics, complete diagrams for the plans generated and their accompanying QUEST diagrams, and a partial trace generated during the creation of the Aladdin narrative.

### A.1 Planning Problem Specification for the Study

The POCL planning algorithm used in the evaluation study and our implementation of the IPOCL algorithm use PDDL-like formulations. A *problem* describes the initial world state and the goal situation. The operator library contains operator schemata. In the evaluation study, the POCL planning algorithm and IPOCL algorithm were given the same inputs. Note however that some parts of the inputs are not used by the POCL algorithm.

The following propositions define the initial state:

| | | |
|---|---|---|
| character(aladdin) | character(jasmine) | character(genie) |
| male(aladdin) | female(jasmine) | monster(genie) |
| knight(aladdin) | at(jasmine, castle) | genie(genie) |
| at(aladdin, castle) | alive(jasmine) | in(genie, lamp) |
| alive(aladdin) | single(jasmine) | confined(genie) |
| single(aladdin) | beautiful(jasmine) | alive(genie) |
| loyal-to(aladdin, jafar) | character(dragon) | scary(genie) |
| character(jafar) | monster(dragon) | place(castle) |
| male(jafar) | dragon(dragon) | place(mountain) |
| king(jafar) | at(dragon, mountain) | thing(lamp) |
| at(jafar, castle) | alive(dragon) | magic-lamp(lamp) |
| alive(jafar) | scary(dragon) | has(dragon, lamp) |
| single(jafar) | | |

The following propositions define the outcome situation:

| | |
|---|---|
| married-to(jafar, jasmine) | ¬alive(genie) |

The following action schemata were provided in the operator library for the evaluation study. Note the deviations from conventional PDDL. *Constraints* indicate immutable propositions that must always be true. Constraints function like preconditions except that they can only be satisfied by the initial state and no operators can negate a proposition that is used as a constraint in an operator schema. The *actors* slot lists the parameters that refer to the actors that intend the operation; we do not assume that the first parameter is always the intentional actor. Further, when there is more than one actor listed, the operator is a *joint operation*, meaning that the operator can only be accomplished by that many actors working as a team and that all actors intend one of the effects of the operator. When *happening* is true, the operator is a is allowed to remain an orphan. Some operators have effects of the form `intends(?x, ?c)` indicating that the effect of one of the characters bound to `?x` has an intention to achieve the literal bound to `?c`. There are no operators that have preconditions of that form; intention propositions are exclusively used by the IPOCL algorithm implementation.

Action: **travel** (?traveller, ?from, ?dest)
    actors: ?traveller
    constraints: character(?traveller), place(?from), place(?dest)
    precondition: at(?traveller, ?from), alive(?traveller), ?from≠?dest
    effect: ¬at(?traveller, ?from), at(?traveller, ?dest)

Action: **slay** (?slayer, ?monster, ?place)
    actors: ?slayer
    constraints: knight(?slayer), monster(?monster), place(?place)
    precondition: at(?slayer, ?place), at(?monster, ?place), alive(?slayer), alive(?monster)
    effect: ¬alive(?monster)

Action: **pillage** (?pillager, ?body, ?thing, ?place)
    actors: ?pillager
    constraints: character(?pillager), character(?body), thing(?thing), place(?place)
    precondition: at(?pillager, ?place), at(?body, ?place), has(?body, ?thing),
                ¬alive(?body), alive(?pillager), ?pillager≠?body
    effect: ¬has(?body, ?thing), has(?pillager, ?thing)

Action: **give** (?giver, ?givee, ?thing, ?place)
    actors: ?giver
    constraints: character(?giver), character(?givee), thing(?thing), place(?place)
    precondition: at(?giver, ?place), at(?givee, ?place), has(?giver, ?thing),
                alive(?giver), alive(?givee), ?giver≠?givee
    effect: ¬has(?giver, ?thing), has(?givee, ?thing)

Action: **summon** (?char, ?genie, ?lamp, ?place)
    actors: ?char
    constraints: character(?char), genie(?genie), magic-lamp(?lamp), place(?place)
    precondition: at(?char, ?place), has(?char, ?lamp), in(?genie, ?lamp),
                alive(?char), alive(?genie), ?char≠?genie
    effect: at(?genie, ?place), ¬in(?genie, ?lamp), ¬confined(?genie), controls(?char, ?genie, ?lamp)

Action: **love-spell** (?genie, ?target, ?lover)
    actors: ?genie
    constraints: genie(?genie), character(?target), character(?lover)
    precondition: ¬confined(?genie), ¬loves(?target, ?lover), alive(?genie), alive(?target), alive(?lover),
                ?genie≠?target, ?genie≠?lover, ?target≠?lover
    effect: loves(?target, ?lover), intends(?target, married-to(?target, ?lover))

Action: **marry** (?groom, ?bride, ?place)
    actors: ?groom, ?bride
    constraints: male(?groom), female(?bride), place(?place)
    precondition: at(?groom, ?place), at(?bride, ?place), loves(?groom, ?bride), loves(?bride, ?groom),
                alive(?groom), alive(?bride)
    effect: married(?groom), married(?bride), ¬single(?groom), ¬single(?bride),
                married-to(?groom, ?bride), married-to(?bride, ?groom)

```
Action: fall-in-love (?male, ?female, ?place)
    actors: ?male
    happening: t
    constraints: male(?male), female(?female), place(?place)
    precondition: at(?male, ?place), at(?female, ?place), single(?male), alive(?male), alive(?female),
                  ¬loves(?male, ?female), ¬loves(?female, ?male), beautiful(?female)
    effect: loves(?male, ?female), intends(?male, married-to(?male, ?female))

Action: order (?king, ?knight, ?place, ?objective)
    actors: ?king
    constraints: king(?king), knight(?knight), place(?place)
    precondition: at(?king, ?place), at(?knight, ?place), alive(?king), alive(?knight),
                  loyal-to(?knight, ?king)
    effect: intends(?knight, ?objective)

Action: command (?char, ?genie, ?lamp, ?objective)
    actors: ?char
    constraints: character(?char), genie(?genie), magic-lamp(?lamp)
    precondition: has(?char, ?lamp), controls(?char, ?genie, ?lamp), alive(?char), alive(?genie),
                  ?char≠?genie
    effect: intends(?genie, ?objective)

Action: appear-threatening (?monster, ?char, ?place)
    actors: ?monster
    happening: t
    constraints: monster(?monster), character(?char), place(?place)
    precondition: at(?monster, ?place), at(?char, ?place), scary(?monster), ?monster≠?char
    effect: intends(?char, ¬alive(?monster))
```

As mentioned in Section 4.5, we required domain-dependent and domain-independent heuristics to generate the example fabula shown in Figure 13 (the plan structure diagram is shown in Figure 15). In IPOCL, heuristics evaluate a plan node and return an integer such that solutions are evaluated to 0 and the higher the number the farther the plan node is from being a solution. We used two heuristic functions whose return values were added together.

- *Domain-independent* heuristic

  - 1 for each action

  - 1 for each flaw

  - 10 for each frame of commitment if there is more than one frame per character

  - 1000 for orphans to be performed by characters for which there are no frames of commitment

- *Domain-dependent* heuristic

  - 5000 for repeat actions

  - 5000 for frames of commitment for character-goal combinations that are not on the following lists:

* Aladdin intends `has(king, lamp)`, `¬alive(genie)`, `¬alive(dragon)`, `has(hero, lamp)`, or `married-to(hero, jasmine)`
* Jafar intends `married-to(jafar, jasmine)`
* Jasmine intends `married-to(jasmine, jafar)`, or `married-to(jasmine, jafar)`
* Genie intends `loves(jasmine, jafar)`, `loves(jafar, jasmine)`, `loves(aladdin, jasmine)`, or `loves(jasmine, jafar)`

  − 1000 if the action `marry` is not associated with two frames of commitment

## A.2 Plan Diagrams and QUEST Structures for the Study

The subsequent figures show the plan diagrams and corresponding QUEST structures for fabula generated for the evaluation.

* Figure 14: a fabula plan automatically generated by a POCL planning algorithm for the POCL condition of the evaluation study.

* Figure 15: a fabula plan automatically generated by our IPOCL algorithm implementation for the IPOCL condition on the evaluation study.

* Figure 16: the QUEST structure corresponding to the fabula plan for the POCL condition of the evaluation study.

* Figure 17: the QUEST structure corresponding to the fabula plan for the IPOCL condition of the evaluation study.

In Figures 14 and 15, solid boxes represent actions in the plan structure (where the last action is the *goal step*). Solid arrows represent causal links and the associated text is both an effect of the preceding action and a precondition of the successive action. Dashed arrows are temporal constraints found by the planning algorithm, indicating a necessary temporally ordering between two actions added due to promotion or demotion strategies for resolving causal threats. In Figure 15, ovals are frames of commitment and horizontal dashed lines represent membership of actions in frames of commitment.

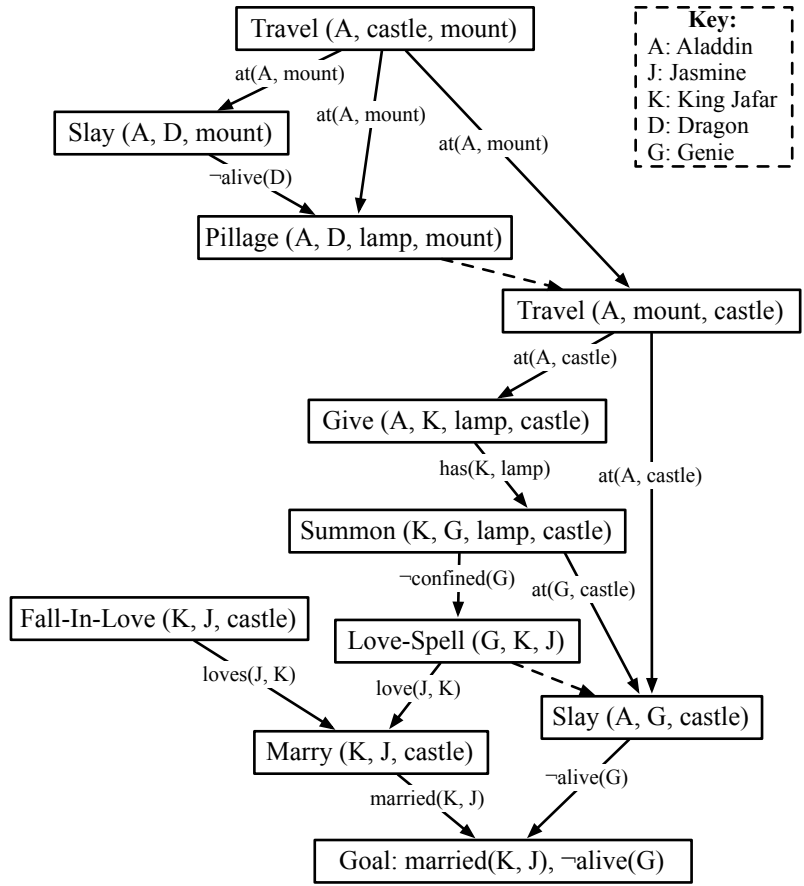

Figure 14: Fabula plan representation of the story used in the POCL condition of the evaluation study.

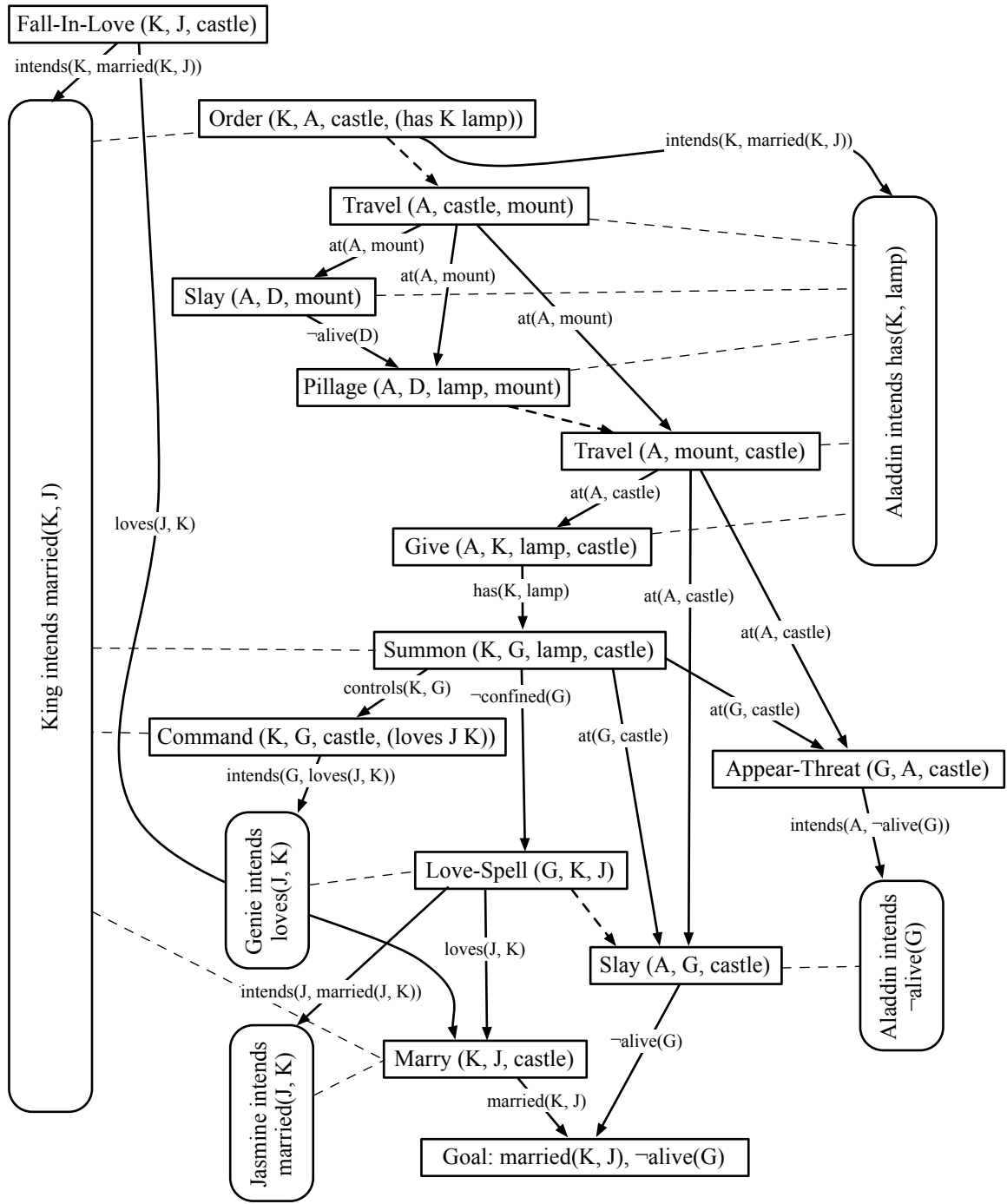

Figure 15: Fabula plan representation of the story used in the experimental IPOCL of the evaluation study.

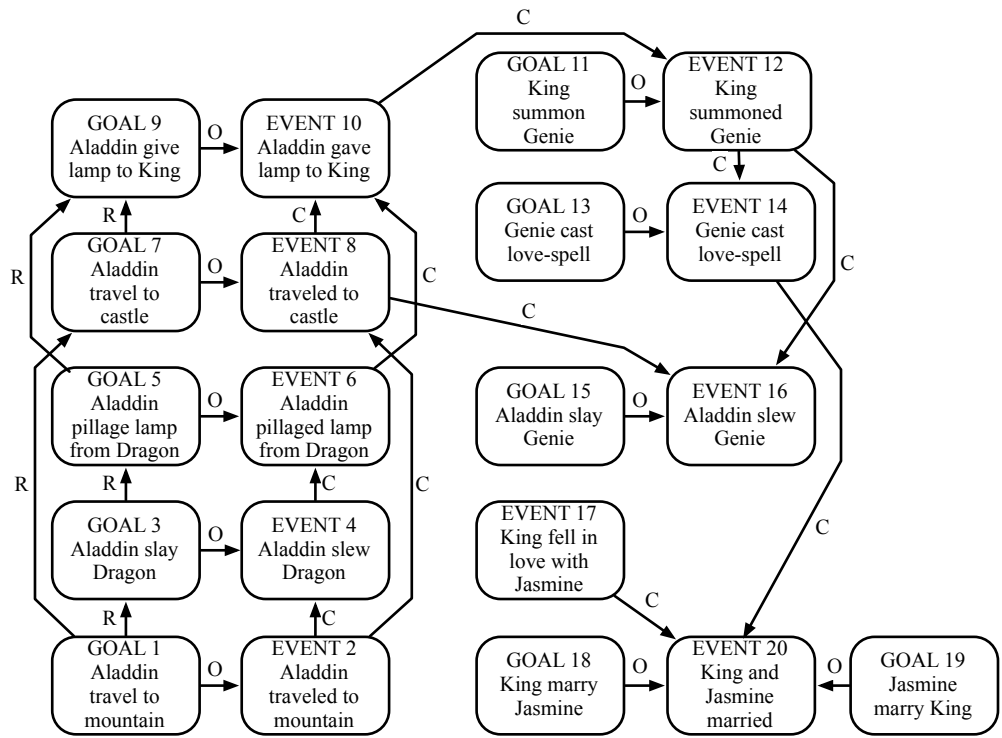

Figure 16: The QUEST knowledge structure for the story plan from the POCL condition.

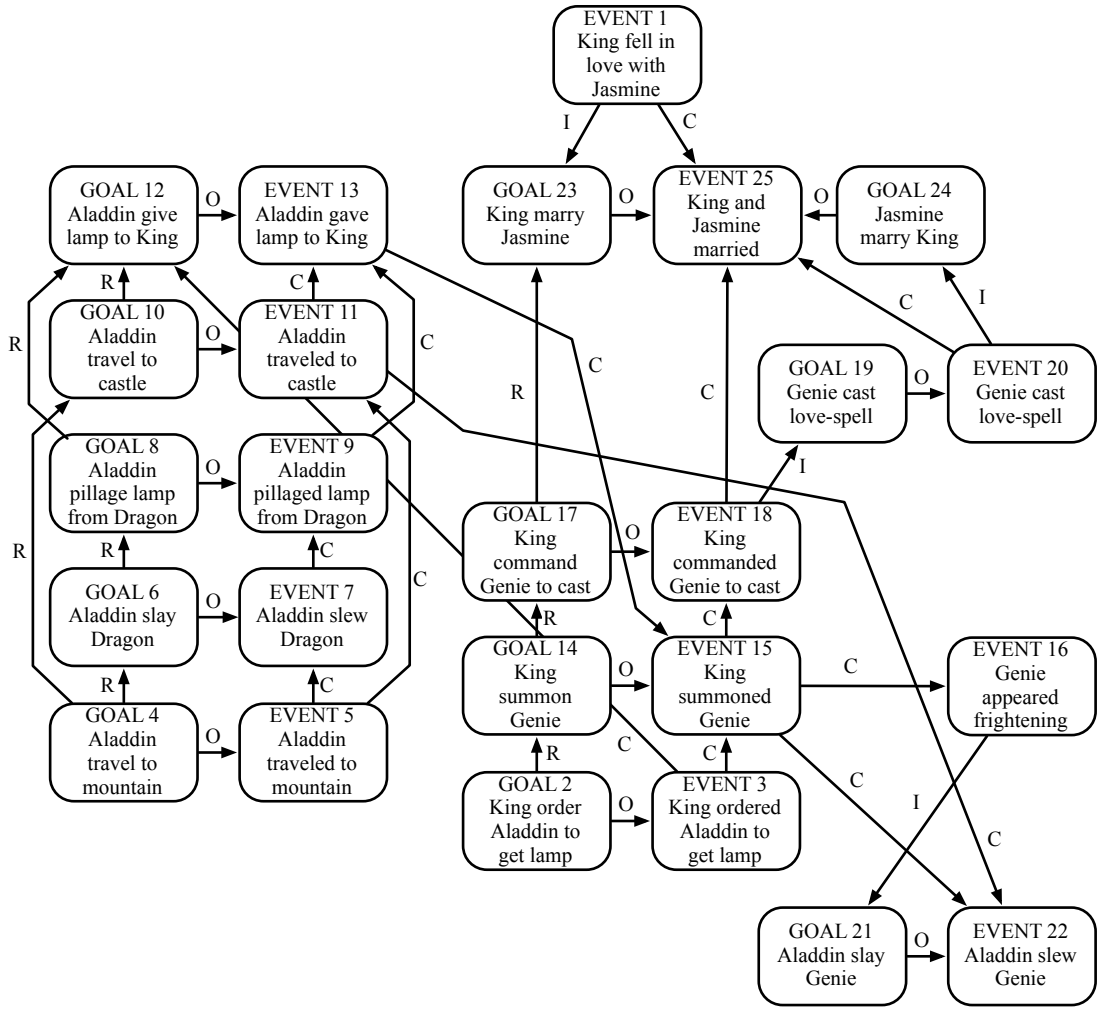

Figure 17: The QUEST knowledge structure for the story plan from the IPOCL condition.

## A.3 Trace

The following is a portion of a trace generated by IPOCL initialized with the above fabula planning problem specification and heuristic. The trace focuses on the generation of some of the nodes in the plan space that contribute to the final solution (shown graphically in Figure 15).

plan0
  reason: initial plan
  now working on: open condition `married-to(jafar, jasmine)` on step goal
  children: 98 (visited 2; selecting 9)

plan 9
  reason: created new step 1: `marry(jafar, jasmine, castle)` to solve `married-to(jafar, jasmine)`
  now working on: open motivation `intends(jafar, married-to(jafar, jasmine))` on frame 2
  children: 9 (visited 2; selecting 105)

plan 105
  reason: created new step 2: `fall-in-love(jafar, jasmine, castle)` to solve
    `intends(jafar, married-to(jafar, jasmine))`
  now working on: open motivation `intends(jasmine, married-to(jasmine, jafar))` on frame 1
  children: 3 (visited 1; selecting 122)

plan 122
  reason: created new step 3: `love-spell(genie, jasmine, jafar)` to solve
    `intends(jasmine, married-to(jasmine, jafar))`
  now working on: open motivation `intends(genie, loves(jasmine, jafar))` on frame 3
  children: 8 (visited 2; selecting 141)

plan 141
  reason: created new step 4: `command(jafar, genie, lamp, loves(jasmine, jafar))` to solve
    `intends(genie, loves(jasmine, jafar))`
  now working on: open condition `alive(genie)` on step 4
  children: 1 (visited 1; selecting 197)

...

plan 591
  reason: created new step 6: `give(aladdin, jafar, lamp, castle)` to solve `has(jafar, lamp)`
  now working on: open motivation `intends(aladdin, has(jafar, lamp))` on frame 4
  children: 4 (visited 2; selecting 4675)

plan 4675
  reason: created new step 7: `order(jafar, aladdin, castle, has(jafar, lamp))` to solve
    `intends(aladdin, has(jafar, lamp))`
  now working on: open condition `loyal-to(aladdin, jafar)` on step 7
  children: 1 (visited 1; selecting 21578)

...

plan 21597
  reason: created new step 8: `pillage(aladdin, dragon, lamp, mountain)` to solve `has(aladdin, lamp)`
  now working on: open condition `alive(aladdin)` on step 8
  children: 1 (visited 1; selecting 21653)

...

plan 1398116
  reason: created new step 12: `slay(aladdin, genie, castle)` to solve ¬`alive(genie)`
  now working on: causal threat on `alive(genie)` between 0 and 3, clobbered by step 12
  children: 1 (visited 1; selecting 1398282)

...

plan 1398289
  reason: created new step 13: `appear-threatening(genie, aladdin, mountain)` to solve
    `intends(aladdin, ¬alive(genie))`
  now working on: open condition `scary(genie)` on step 13
  children: 1 (visited 1; selecting 1398304)

...

plan 1398364
  reason: adoption of step 4 by frame 2: jafar intends `married-to(jafar, jasmine)`
  now working on: intent flaw for aladdin, to possibly link step 8 to frame 4: aladdin intends
    `has(jafar, lamp)`
  children: 2 (visited 2; selecting 1398368)

plan 1398368
  reason: adoption of step 8 by frame 4: aladdin intends `has(jafar, lamp)`
  now working on: intent flaw for aladdin, to possibly link step 10 to frame 4: aladdin intends
    `has(jafar, lamp)`
  children: 2 (visited 2; selecting: 1398376)

...

plan 1398384
  reason: no adoption of step 2 by frame 2: jafar intends `married-to(jafar, jasmine)`
  now working on: intent flaw for jasmine, to possibly link step 12 to frame 1: jasmine intends
    `married-to(jasmine, jafar)`
  children: 2 (visited 2; selecting 1398400)

...

plan 1398576
  reason: adoption of step 7 by frame 2: jafar intends `married-to(jafar, jasmine)`
  now working on: intent flaw for aladdin, to possibly link step 9 to frame 4: aladdin intends
    `has(jafar, lamp)`
  children: 2 (visited 1; selecting 1398640)

plan 1398640
  reason: adoption of step 9 by frame 4: aladdin intends `has(jafar, lamp)`
  solution found